\documentclass[conference]{IEEEtran}
\IEEEoverridecommandlockouts
\usepackage{cite}
\usepackage{amsmath,amssymb,amsfonts}
\usepackage{algorithmic}
\usepackage{graphicx}
\usepackage{subfigure}
\usepackage{textcomp}
\usepackage{hyperref}

\usepackage[hyphenbreaks]{breakurl}
\usepackage{xurl}
\def\BibTeX{{\rm B\kern-.05em{\sc i\kern-.025em b}\kern-.08em
    T\kern-.1667em\lower.7ex\hbox{E}\kern-.125emX}}
\usepackage{cleveref}
\usepackage{booktabs}
    
\usepackage{xcolor}

 \usepackage{amssymb} 
 \usepackage{pifont}

\usepackage{listings}
\usepackage{fancyvrb}
\usepackage{framed}
\usepackage{color}
\usepackage{pgfplots}
\usepackage{tikz}
\usetikzlibrary{positioning, decorations.pathreplacing, arrows.meta}
\usepackage{orcidlink}
\usepackage{tabularx}
\definecolor{light-gray}{gray}{0.95}
\hypersetup{
   breaklinks=true,   
   colorlinks=true,   
   pdfusetitle=true,  
}
 
\begin{document}

\title{Blessing or curse? A survey on the Impact of Generative AI on Fake News}

\author{
\IEEEauthorblockN{Alexander Loth\orcidlink{0009-0003-9327-6865}}
\IEEEauthorblockA{\textit{Microsoft} \\
Alexander.Loth@microsoft.com}
\and
\IEEEauthorblockN{Martin Kappes}
\IEEEauthorblockA{\textit{Frankfurt University of Applied Sciences} \\
kappes@fb2.fra-uas.de}
\and
\IEEEauthorblockN{Marc-Oliver Pahl\orcidlink{0000-0001-5241-3809}}
\IEEEauthorblockA{\textit{IMT Atlantique, UMR IRISA, Chaire Cyber CNI} \\
marc-oliver.pahl@imt-atlantique.fr}
}

\maketitle

\begin{abstract}
Fake news significantly influence our society.
They impact consumers, voters, and many other societal groups.
While Fake News exist for a centuries, Generative AI brings fake news on a new level.
It is now possible to automate the creation of masses of high-quality individually targeted Fake News.
On the other end, Generative AI can also help detecting Fake News.
Both fields are young but developing fast.
This survey provides a comprehensive examination of the research and practical use of Generative AI for Fake News detection and creation in 2024.
Following the Structured Literature Survey approach, the paper synthesizes current results in the following topic clusters 
1) enabling technologies,
2) creation of Fake News,
3) case study social media as most relevant distribution channel, 
4) detection of Fake News,
and
5) deepfakes as upcoming technology.
The article also identifies current challenges and open issues.
\end{abstract}

\begin{IEEEkeywords}
Generative Artificial Intelligence, Generative AI, GenAI, Fake News Detection, Deep Learning, Information Dissemination, Social Media, Natural Language Processing, NLP, Ethical AI, Misinformation, Content Creation, Synthetic Media, Information Security, Misinformation Mitigation.
\end{IEEEkeywords}

\section{Introduction}
\label{introduction}







The development of information dissemination has a long history of influencing public perception and political landscapes.
In the era of Stalin, photo manipulation was employed to erase political rivals from historical records, altering public perception of history.
Goebbels, the Reich Minister of Propaganda in Nazi Germany, masterfully used propaganda to control public opinion and further the Nazi agenda.

Cambridge Analytica, a British political consulting firm, became widely known for its role in the 2016 United States presidential election and the Brexit referendum\cite{berghel_malice_2018}.
Their case shows the immense influential potential of individually-targeted information.
The company specialized in data mining, data analysis, and strategic communication during electoral processes.
The Cambridge Analytica scandal used data analytics in influencing voter behavior through targeted, at the time, hand crafted political advertising on social media.
Cambridge Analytica's techniques were used in various political campaigns, including the 2016 US presidential election and the Brexit referendum.
The company claimed that its data-driven approach was highly effective in shaping political outcomes.

Generative AI significantly transforms our society\cite{loth_openai_2023}.
This technology can automate the creation of realistic images, videos, and texts, making it increasingly difficult to distinguish between real and artificial content.
The generation comes at almost no cost, enabling the massive creation of individually-targeted information.
The sophistication of Generative AI raises unprecedented challenges in the realms of news, media, and information integrity.

Fake News specifically refers to fabricated information that mimics news media content in form but not in organizational process or intent, disinformation is false information deliberately spread to deceive people, and misinformation is false information spread without an intention to mislead\cite{shu_fake_2017, wardle2017information}.
The proliferation of Fake News represents a significant challenge, necessitating advanced solutions for detection and mitigation\cite{sharma_combating_2019}.
In this context, Generative Artificial Intelligence (GenAI) has emerged as a dual-use technology capable of both generating sophisticated fake content and detecting it.

Surveys have started exploring various facets of Fake News and Generative AI, highlighting the technical challenges and the development of both identification and mitigation techniques\cite{sharma_combating_2019}.
This survey fills a gap by systematically exploring how Generative AI technologies are applied to both propagate and combat Fake News.
The rapid evolution of Generative AI techniques, particularly in Natural Language Processing (NLP) and deep learning, underscores the urgency for an updated analysis of this domain.


This survey addresses the interplay between Generative AI and Fake News:
\begin{enumerate}
    \item How is Generative AI used in the context of Fake News creation and detection?
    \item How is the domain of Generative AI structured in the context of Fake News?
    \item How to cluster existing research systematically?
    \item What are the relevant venues and the active research groups worldwide driving this field?
    \item What are currently open issues and what is possible future work?
\end{enumerate}

This survey delineates several key contributions to the literature on Generative AI and Fake News, marking a novel departure from existing surveys:
\begin{enumerate}
    \item Coming from a technological perspective, this work structures the Generative AI domain used and usable for Fake News.
    \item From a research perspective, this work presents and clusters existing research streams and practical work and offers a categorization.
    \item Also from a research perspective, this work structures the field by identifying the current publication hot spots (venues where papers were published).
    \item In the same systematic, it identifies the research hotspots with the active groups.
    \item Finally, this survey highlights open issues and extrapolates relevant future research directions, offering a research roadmap.
%
\end{enumerate}


This survey is structured as follows:
\Cref{related} provides the context of this research, showing that it is missing.
\Cref{domain} introduces the key concepts and aspects of Fake News and Generative AI.
\Cref{methodology} details the methodological applied for this work.
\Cref{overview} provides a comprehensive in-depth review of relevant recent research.
\Cref{findings} dissects the evolution of the research topic and presents the current central publication venues and research groups.
\Cref{state} synthesizes the findings from the survey, distilling the essence of current research and its implications for the future.



\section{Related Work}
\label{related}


This survey identifies a gap in the intersection of Generative AI and Fake News generation and detection.
Most studies focus on utilizing Natural Language Processing and machine learning.
The generative capabilities of AI in creating Fake News receive less attention.
While advancements in Generative AI are documented, their direct impact on Fake News remains secondary.
Social media's role in misinformation is well-studied, yet the potential of Generative AI in content creation is often overlooked.
Interdisciplinary approaches to Fake News are common, but the specific impact of Generative AI is less explored.
Deepfake research focuses on specific applications without a broader view of Generative AI in Fake News.
This survey fills these gaps, highlighting Generative AI's role in both generating and detecting misinformation.


\subsection{Presentation and Analysis of Related Works}

While numerous components already exist in the field, this research contributes a novel perspective by focusing on Generative AI.
The related work is structured in five sub-sections (as described in \Cref{overview}):

\begin{itemize}
    \item \textbf{Enabling technologies}
    \item \textbf{Creation of Fake News}
    \item \textbf{Case study social media as most relevant distribution channel}
    \item \textbf{Detection of Fake News}
    \item \textbf{Deepfakes as upcoming technology}
\end{itemize}

\subsubsection{Enabling technologies}

\textbf{Oshikawa et al. (2020)} delve into the application of Natural Language Processing (NLP) for Fake News detection, highlighting the critical role of NLP in automating the identification of false information online\cite{oshikawa_survey_2020}. Similar to our survey, Oshikawa et al. emphasize the importance of technological solutions in combating Fake News. Unlike their work, we focus on the generative aspect of AI, exploring how these technologies not only detect but also contribute to the creation of Fake News.

\textbf{Villela et al. (2023)} analyze machine learning algorithms used in Fake News detection. It aligns with our survey in terms of the technological approach to combating Fake News but does not delve into the specific role of Generative AI for creating Fake News\cite{villela_fake_2023}.

\textbf{Leo (2019)} examines machine learning applications in banking risk management. Although not directly related to our survey's focus, it provides insights into machine learning applications in a different field, showing the versatility of these technologies\cite{leo_machine_2019}.

\subsubsection{Creation of Fake News}

\textbf{Cao et al. (2023)} provide an extensive historical overview of Generative AI, including ChatGPT and DALL-E-2. They focus on the efficiency and accessibility of AI in content creation. The survey's emphasis on the technological evolution and capability of Generative AI models aligns with our focus on AI's role in content generation. However, it does not specifically focus on the Fake News aspect, which is central to our survey\cite{cao_comprehensive_2023}.

\textbf{Kalyan \& Ammus (2021)} delve into transformer-based pretrained models in NLP, crucial for understanding modern Generative AI. It provides a technical foundation that is relevant to our survey but does not directly address the application of these technologies in generating or detecting Fake News\cite{kalyan_ammus_2021}.

\subsubsection{Case study social media as most relevant distribution channel}

\textbf{Melchior et al. (2023)} investigate the motivations behind sharing Fake News on social media, exploring both intrinsic and extrinsic factors. While it provides insights into the psychology of Fake News distribution, the article lacks the emphasis on Generative AI's role in content creation, which is a key area of our survey\cite{melchior_systematic_2023}.

\textbf{Shu et al. (2017)} review Fake News detection on social media, emphasizing the challenges and unique characteristics of social media platforms. It intersects with our survey in the context of Fake News but only covers aspects of Generative AI in this process\cite{shu_fake_2017}.

\subsubsection{Detection of Fake News}

\textbf{Zhou et al. (2020)} explore Fake News detection from various perspectives, including the content's falsehood, style, propagation patterns, and source-credibility. It encourages interdisciplinary research on Fake News. While it shares our focus on Fake News, our survey specifically examines the intersection with Generative AI, which is not the focus of Zhou et al.\cite{zhou_survey_2020}.

\textbf{Bondielli and Marcelloni (2019)} investigate various techniques for detecting Fake News and rumors, underscoring the significance of classification, data mining, and machine learning methods\cite{bondielli_survey_2019}. While Bondielli and Marcelloni provide an extensive overview of detection methodologies, our survey uniquely integrates the generative capabilities of AI, examining both the challenges and opportunities they present in the context of misinformation.

\textbf{Sharma et al. (2019)} present an in-depth exploration of identification and mitigation techniques for combating Fake News\cite{sharma_combating_2019}. Their work is akin to ours in its multidisciplinary approach to addressing Fake News. However, our survey distinguishes itself by specifically analyzing the impact of Generative AI in both the propagation and detection of Fake News, filling a gap they have not explored.

\subsubsection{Deepfakes as upcoming technology}

\textbf{Khanjani (2023)} focuses on audio deepfakes, discussing their creation, detection, and impact. While it explores a specific application of Generative AI, its emphasis on audio content differentiates it from our survey, which has a broader focus on Fake News encompassing text and visual content\cite{khanjani_audio_2023}.

These works underscore the advancements and challenges in the domains of Generative AI and Fake News.
The present study differs by focusing on the synergy of these domains, proposing a unified approach to understanding and mitigating Fake News through Generative AI.

\section{The Domain}
\label{domain}




The term ``Generative Artificial Intelligence (AI)'' refers to AI technologies designed to produce new and original content. 
This content includes text, images, audio, and other media forms, often matching the sophistication and authenticity of human-generated output.
Models like GPT-4, with their human-like text generation, are able to pass the Turing test\cite{biever_chatgpt_2023}.

Advanced machine learning models, such as Generative Adversarial Networks (GANs), transformers, and variational autoencoders, enable this capability. 
By harnessing extensive datasets, these models learn to generate novel instances that mimic the patterns they have acquired\cite{cao_comprehensive_2023,gozalo-brizuela_chatgpt_2023}.

As Generative AI evolves, it simultaneously enhances the ability to create synthetic content and offers tools for its detection.
This dual-use nature highlights the need for technologies that foster trust and authenticity. 
Sandner describes blockchain as the ``digital notary of the 21st century—a guardian of truth in an age of uncertainty''\cite{sandner_blockchain_2024}. 
Blockchain technology provides immutable proof of content authenticity and ensures transparency across decentralized networks, offering a promising solution to verify origins and detect tampering in digital content.
This approach strengthens trust and accountability, helping counter misinformation and securing data integrity.

The synthesis of recent advancements in Generative AI and its application to the generation and detection of Fake News reveals a rapidly evolving field marked by interdisciplinary contributions. From the foundational work by Devlin et al. (2018) on BERT, which revolutionized natural language processing (NLP) through deep bidirectional transformers, to the innovative detection models like \textit{exBAKE} by Jwa et al. (2019) and the application of transformers in identifying automatically generated headlines by Maronikolakis et al. (2021), the landscape of Fake News detection has significantly expanded\cite{devlin_bert_2018,jwa_exbake_2019,maronikolakis_identifying_2021}.

Cao et al. (2020) and Liu et al. (2019) extend the discussion to multimedia content and short Fake News stories, respectively, showcasing the breadth of content that now falls under the purview of Fake News detection efforts\cite{cao_exploring_2020,liu_two-stage_2019}. The pandemic has further underscored the critical need for timely and accurate detection mechanisms, as evidenced by Karnyoto et al. (2022) and Novo (2021), who focus on COVID-19 misinformation\cite{karnyoto_transfer_2022,novo_misinfobot_2021}.

Emerging concerns around synthetic media, or "deepfakes"—highly realistic AI-generated audiovisual content intended to deceive by impersonating real individuals—have prompted researchers like Bansal et al. (2023) and Botha and Pieterse (2020) to develop detection methods that address the nuanced challenges posed by this content\cite{bansal_deepfake_2023,botha_fake_2020}.
Meanwhile, the work of Wang et al. (2020) and Cardenuto et al. (2023) highlights the importance of distinguishing between genuine and artificially created images and realities\cite{wang_fakespotter_2020,cardenuto_age_2023}.

The role of AI in both facilitating and combating the spread of Fake News introduces a paradox that researchers must navigate. While generative AI models have the potential to create highly realistic and misleading content, as explored by Mosallanezhad et al. (2020) and Cocchi et al. (2023), they also offer the tools necessary for the development of sophisticated detection algorithms\cite{mosallanezhad_topic-preserving_2020,cocchi_unveiling_2023}.

The inception of generative models can be traced back to the 1950s, marked by the development of Hidden Markov Models (HMMs) and Gaussian Mixture Models (GMMs). 
However, the introduction of GANs and the progression of models such as BERT, GPT, and DALL-E represented transformative milestones in the field\cite{cao_comprehensive_2023}. 
These models have been instrumental in producing content that is not only realistic but also of high quality, paving the way for their extensive application across various industries.

Alongside its rise, Generative AI has sparked significant concerns regarding its potential misuse, particularly in the generation of Fake News. 
Fake News, which is characterized by the deliberate presentation of misleading or false information as legitimate news, has become an increasingly problematic issue. 
This is especially true in the context of social media platforms that enable rapid and widespread dissemination of information\cite{allcott_social_2017,lazer_science_2018,shu_fake_2017}. 
The capacity of Generative AI to create convincingly authentic text and media intensifies this problem, as it blurs the line between genuine and AI-generated content, making it progressively difficult to differentiate between the two.

\begin{figure*}[ht!]
\centering
\resizebox{\columnwidth}{!}
{
\begin{tikzpicture}[>=Stealth,shorten >=1pt,auto,node distance=1.5cm,
  thick,main node/.style={rectangle, rounded corners,fill=blue!20,draw},
  sub node/.style={rectangle, rounded corners,fill=green!20,draw,align=center},
  inner node/.style={rectangle, rounded corners,fill=yellow!20,draw,align=center},
  legend node/.style={font=\footnotesize, rectangle, fill=white,draw=none,align=left}]

  \node[main node] (1) {Generative AI};
  \node[main node] (2) [above left = of 1] {(1) Fake News Creation};
  \node[main node] (3) [above right = of 1] {(2) Fake News Detection};
  \node[main node] (8) [below left = of 1] {(3) Mitigation Strategies};
  \node[main node] (9) [below right = of 1] {(4) Ethical Considerations};

  \node[sub node] (4) [above = of 2] {Text\\Generation};
  \node[sub node] (5) [left = of 4] {Image\\Synthesis};
  \node[sub node] (19) [right = of 4] {Audio\\Generation};
  \node[sub node] (20) [above = of 5] {Video\\Generation};

  \node[sub node] (7) [above = of 3] {Social Media\\Analysis};
  \node[sub node] (6) [right = of 7] {Content\\Verification};
  \node[sub node] (21) [below = of 6] {Crowd\\Sourcing};

  \node[inner node] (10) [right = of 1] {Autoencoders};
  \node[inner node] (11) [right = of 10] {VAEs};
  \node[inner node] (12) [left = of 1] {Transformers};
  \node[inner node] (17) [left = of 12] {GPTs};
  \node[inner node] (18) [above = of 1] {GANs};

  \node[sub node] (13) [below = of 8] {Public\\Awareness};
  \node[sub node] (14) [left = of 13] {Regulatory\\Policies};

  \node[sub node] (15) [below = of 9] {Privacy\\Concerns};
  \node[sub node] (16) [right = of 15] {Bias \&\\Fairness};

  \node[legend node,anchor=south] at (current bounding box.south) (legend) {%
    \textbf{Abbreviations:}\\
    GPT: Generative Pre-trained Transformer\\
    VAE: Variational Autoencoder\\
    GAN: Generative Adversarial Network
  };

  \path[every node/.style={font=\sffamily\small}]
    (1) edge node [left] {} (2)
        edge node [right] {} (3)
        edge node [left] {} (10)
        edge node [right] {} (12)
        edge node [left] {} (8)
        edge node [right] {} (9)
        edge node [right] {} (18)
    (2) edge node [left] {} (4)
        edge node [left] {} (5)
        edge node [right] {} (19)
    (3) edge node [right] {} (6)
        edge node [right] {} (7)
        edge node [right] {} (21)
    (8) edge (13)
        edge (14)
    (9) edge (15)
        edge (16)
    (4) edge [bend left] node[left] {} (6)
    (5) edge [bend left] node[left] {} (6)
    (7) edge [bend right] node[right] {} (6)
    (19) edge [bend left] node[left] {} (6)
    (20) edge [bend left] node[left] {} (6)
    (10) edge node [right] {} (11)
    (12) edge node [right] {} (17)
    (5) edge node [right] {} (20);
\end{tikzpicture}
}
\caption{Structural overview of the domain of Generative AI and its impact on Fake News, illustrating the main themes of creation and detection, along with their subtopics and interrelations. Blue indicates the main themes of the domain; green highlights the subtopics related to these main themes; and yellow denotes key technologies and methodologies central to understanding and addressing the challenges in this domain.}
\label{fig:generative-ai-fake-news-structure}
\end{figure*}
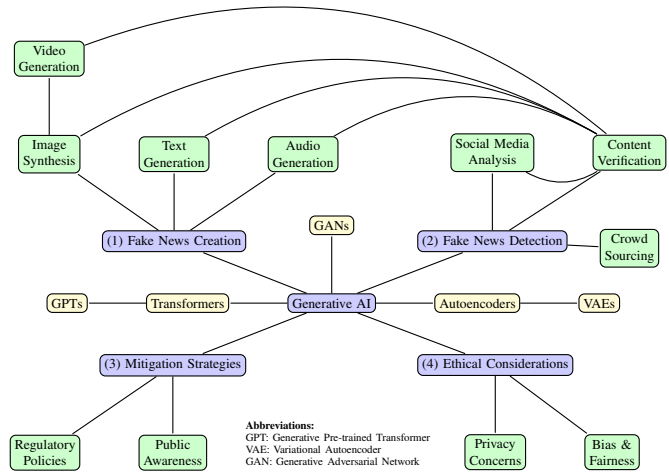

The domain of Fake News in the context of Generative AI encompasses the intersection of misinformation dissemination and advanced computational techniques designed to generate or detect such content.
Generative AI refers to the subset of AI technologies capable of creating content, including text, images, and videos, that is indistinguishable from content created by humans. 
This technology has profound implications for the production and dissemination of news, particularly concerning the authenticity and trustworthiness of online content. 

The advent of Generative AI has ushered in new methods for creating highly realistic and convincing Fake News, a phenomenon that poses significant challenges to information tendencies on digital platforms\cite{zhou_network-based_2019}. 
Fake News, defined as news articles and stories that are factually incorrect yet designed to appear credible, has been identified as a major societal problem.
It influences public opinion, affecting electoral processes, and undermining trust in media institutions\cite{lazer_science_2018}. 

The impact of Generative AI on Fake News is multifaceted, affecting both the creation of deceptive content and the development of detection mechanisms. 
On one hand, Generative AI models, such as GPT-3 and DALL-E, have demonstrated the ability to generate coherent and persuasive text and imagery, raising concerns about their potential misuse for generating disinformation. 
On the other hand, these advances also offer promising avenues for combating Fake News, through the development of AI-driven tools capable of identifying and flagging synthetic content with high accuracy.

The domain's complexity is further amplified by the dynamic nature of social media platforms, where much of today's Fake News is propagated. 
The algorithms governing content dissemination on these platforms play a crucial role in amplifying or mitigating the spread of Fake News, making the study of Generative AI's impact on Fake News an interdisciplinary challenge that spans computer science, communications, and behavioral science\cite{bradshaw_industrialized_2021, shu_combating_2020}.

Given these considerations, the domain of Generative AI and Fake News encompasses not only the technological advancements that facilitate the creation of synthetic content but also the societal implications of these developments, including ethical concerns, regulatory challenges, and the broader effects on democratic discourse. 
This literature review elucidates the current state of research, identifying key trends, challenges, and opportunities for future investigation.

The diagram presented in Figure \ref{fig:generative-ai-fake-news-structure} shows a visual representation of the domain of Generative AI and its multifaceted impact on Fake News.
At the core, Generative AI serves as the foundation, branching out into two principal thematic areas: the creation and detection of Fake News.
The Creation aspect~(1) is further subdivided into methods such as Text Generation, Image Synthesis, Audio Generation, and Video Generation.
These represent the diverse capabilities of Generative AI to produce content that is often indistinguishable from that created by humans.

The Detection branch~(2) is concerned with identifying and verifying the authenticity of content on social media and other platforms, and is categorized into Content Verification, Social Media Analysis, and Crowd Sourcing.
These areas reflect the different strategies used to discern the origins of news items and validate their truthfulness.

Adjacent to these core themes, the diagram also integrates crucial peripheral domains such as mitigation strategies and ethical considerations.
Mitigation strategies~(3) encompass public awareness and regulatory policies, underscoring the societal and governmental responses to the challenges posed by Fake News.
Ethical considerations~(4), including privacy concerns and issues of bias and fairness, reflect the moral implications of using Generative AI in the context of news dissemination.

Connecting these nodes are sub-nodes that represent the enabling technologies of Generative AI, including Autoencoders, GANs (Generative Adversarial Networks), Transformers, GPTs (Generative Pre-trained Transformers), and VAEs (Variational Autoencoders).
These technologies underpin the operations of Generative AI and are integral to both the creation of synthetic content and the development of systems to detect and analyze Fake News.

\subsection{Functioning of Generative AI for Fake News Generation}

Generative AI models synthesize new data by learning from existing datasets. 
They function through deep learning architectures such as GANs, VAEs, and Transformers. 
GANs pit two neural networks against each other to produce new, synthetic instances of data.
GANs generate realistic images and videos to accompany synthetic Fake News stories.

Transformers utilize attention mechanisms to generate coherent sequences of text\cite{vaswani_attention_2023}.
Transformers, like GPT models, are trained on vast corpora of text.
Transformers are able to produce all kind of text, including Fake News\cite{sandrini_generative_2023, ferrara_genai_2024}.

\subsection{Technical Background}
This section provides an overview of the key technologies and concepts foundational to understanding the intersection of Generative AI and Fake News. It introduces the essential definitions and methodologies employed in the survey.

\subsubsection{Generative Artificial Intelligence}
Generative AI refers to a subset of AI technologies designed to create content that mimics real-world data. These models learn to generate new data samples that are indistinguishable from authentic datasets.

\subsubsection{Generative Adversarial Networks (GANs)}
GANs consist of two neural networks, the generator and the discriminator, which are trained simultaneously through adversarial processes. The generator creates data samples aimed at fooling the discriminator, while the discriminator evaluates them against real data, improving both models iteratively\cite{goodfellow_generative_2014}.

\subsubsection{Variational Autoencoders (VAEs)}
VAEs are generative models that use a probabilistic approach to produce data. They learn to encode input data into a latent space and reconstruct it back, ensuring that generated samples adhere to the probability distribution of the input data\cite{kingma_auto-encoding_2022}.

\subsubsection{Transformer Models}
Transformers are a type of neural network architecture designed for processing sequential data, particularly text.
They rely on self-attention mechanisms to weigh the significance of different parts of the input data\cite{vaswani_attention_2023}.
A recent breakthrough in this area is the development of 1-bit Large Language Models (LLMs), such as those introduced by Ma et al. (2024)\cite{ma_era_2024}, which achieve comparable performance to full-precision models with significantly reduced computational costs.

\paragraph{BERT}
Bidirectional Encoder Representations from Transformers (BERT) is a model designed to pre-train deep bidirectional representations from unlabeled text by jointly conditioning on both left and right context in all layers\cite{devlin_bert_2018}.

\paragraph{GPT}
Generative Pre-trained Transformer (GPT) models represent a significant advancement in language processing, capable of generating coherent and contextually relevant text based on a given prompt, which is a user-defined input that guides the model's text generation process.
These models are distinguished by their ability to perform a wide array of language tasks without task-specific training\cite{radford_improving_2018}.

\begin{figure}[htbp]
\centering
\begin{tikzpicture}[>=Stealth, node distance=1cm and 1.5cm, block/.style={draw, rectangle, minimum height=1.5cm, minimum width=3cm}, layer/.style={draw, rectangle, minimum height=0.5cm, minimum width=2cm, fill=gray!20}]

\node[block] (input) {Input Embedding};
\node[layer, below=of input] (pos) {Position Encoding};
\draw[->] (input) -- (pos);

\node[block, below=of pos] (transformer1) {Transformer Block 1};
\draw[->] (pos) -- (transformer1);

\node[block, below=of transformer1] (transformer2) {Transformer Block 2};
\draw[->] (transformer1) -- (transformer2);

\node[block, below=of transformer2] (transformerN) {Transformer Block N};
\draw[->] (transformer2) -- node[midway, fill=white] {$\vdots$} (transformerN);

\node[block, below=of transformerN] (output) {Output Layer};
\draw[->] (transformerN) -- (output);
\end{tikzpicture}

\caption{Schematic representation of the GPT architecture. The model processes input through an embedding layer followed by position encoding to maintain sequence order. This is then passed through multiple transformer blocks, each comprising multi-head self-attention mechanisms and feed-forward neural networks, to generate the output text. The architecture illustrates the flow from input to the generated output.}
\label{fig:gpt_architecture}
\end{figure}
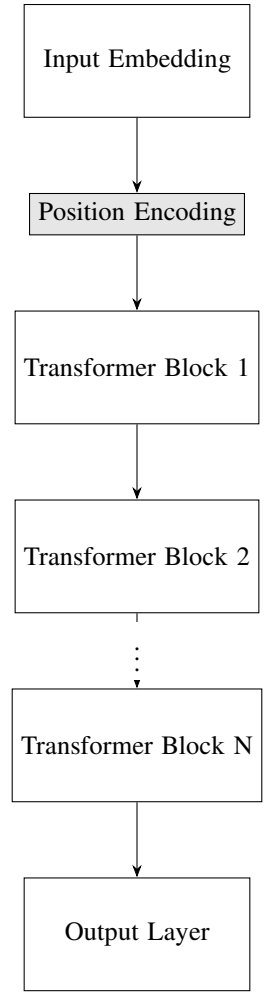

As described in Figure~\ref{fig:gpt_architecture}, the GPT model architecture is designed to capture and generate human-like text by processing input through a series of transformer blocks.
Each block enhances the model's understanding of language context and structure, allowing for the generation of coherent and contextually relevant text.
This mechanism allows the model to simulate various forms of written content, including Fake News, by leveraging learned patterns from extensive data sets.

\section{Methodology}
\label{methodology}


This survey adopts a structured literature review approach. 
It identifies, analyzes, and synthesizes the current research on the use of Generative AI for Fake News. 
The methodology ensures the aggregation of all relevant across diverse fields.

\subsection{Rationale Behind Selection of Specific Works}

Several key factors drive the selection of the specific works presented in this survey.
They are outlined below.

The selected works propose novel methodologies, or present comprehensive overviews that are foundational to understanding the current state of research in Generative AI and Fake News.
The selected literature encompasses a diverse range of perspectives, including technological, psychological, and sociopolitical angles.

Other works, while potentially valuable, are excluded due to reasons such as redundancy, lack of direct relevance to the specific intersection of Generative AI and Fake News, or failure to meet the scholarly standards set for this survey.

\subsection{Search Strategy}

This survey conducts a systematic search across well-established academic databases and search engines. 
The used platforms are:
\begin{itemize}
    \item Web of Science
    \item Google Scholar
    \item IEEE Xplore
    \item ACM Digital Library
    \item PubMed
\end{itemize}

This survey covers publications up to March 2024. 
The search encompasses broad and specific aspects of Generative AI and Fake News, focusing on a curated selection of keywords. 
Keywords target Generative AI and Fake News:
\begin{itemize}
    \item ``Generative AI'' AND ``Fake News''
    \item ``Fake News detection'' AND ``Fake News''
    \item ``AI-generated content'' AND ``Fake News''
    \item ``Deepfakes'' AND ``Fake News''
    \item ``Fake content detection'' AND ``Fake News''
\end{itemize}

\subsection{Inclusion and Exclusion Criteria}

This survey sets specific inclusion criteria to ensure relevance and quality of the literature:
\begin{itemize}
    \item Studies focusing on Generative AI's role in creating, detecting, or mitigating Fake News.
    \item Articles offering empirical evidence or theoretical analysis on Generative AI's impact on misinformation.
    \item Works published in peer-reviewed journals or conference proceedings.
    \item Books that provide a broader perspective for the general audience.
\end{itemize}

Exclusion criteria are:
\begin{itemize}
    \item Studies not directly related to Generative AI and Fake News.
\end{itemize}

\subsection{Selection Process}

An initial search identified 112 articles. 
After removing duplicates, 96 articles were screened by title and abstract. 
This led to 37 articles being excluded for not meeting the inclusion criteria. 
A full-text review of the remaining articles resulted in 59 articles used for this survey.

\subsection{Data Extraction and Synthesis}

For each selected study, this survey extracts:
\begin{itemize}
    \item Publication year,
    \item research objectives,
    \item Generative AI technologies used,
    \item main findings, and
    \item conclusions.
\end{itemize}
This data is synthesized to identify trends, challenges, and gaps, focusing on Generative AI's intersection with Fake News.
See table \ref{tab:topics}.

\subsection{Source types}

This survey categorizes the sources into scientific articles, presentations, and books.
Scientific articles, providing empirical evidence, form the empirical backbone of this analysis.
Presentations, offering insights on current practices, enrich the discourse with real-world applications.
Books, delivering comprehensive views, extend the understanding of AI's societal impact.

\subsubsection{Scientific sources}
Scientific sources lay the foundation for significant advancements in AI technologies, especially in the realms of NLP and deep learning models. 
These advancements are critical for the enhanced detection and analysis of Fake News. 
Published primarily in peer-reviewed journals and conference proceedings, these sources furnish empirical evidence and theoretical frameworks. 
Such frameworks are vital for navigating and mitigating the complex dynamics of misinformation. 
The focus on multimedia content by Cao et al. (2020), alongside the innovative employment of BERT and GPT models by Karnyoto et al. (2022) and Jwa et al. (2019), mirror the evolving methodologies within this domain\cite{cao_exploring_2020,karnyoto_transfer_2022,jwa_exbake_2019}. 
Hence, scientific literature serves as a pivotal cornerstone for future explorations, grounding the discussion on generative AI's dual role in creating and detecting fake content.

\subsubsection{Presentations}
Presentations focus on the practical applications and implications of AI technologies in combating Fake News.
They serve as platforms for discussing cutting-edge methodologies and sharing insights.
The presentation by Schütz et al. (2021) demonstrates the use of pre-trained transformer models for automatic Fake News detection, showcasing practical applications of generative AI\cite{del_bimbo_automatic_2021}.
It emphasizes the efficiency of fine-tuning state-of-the-art AI technologies to distinguish between fact and fiction.
These forums are essential for the real-time exchange of ideas and critiques, fostering a collaborative environment that advances the field.
Loth (2023) adds to this discourse by exploring Generative AI's broader impact on innovation and enhancing human collaboration, underlining the necessity of ethical frameworks in AI development\cite{loth_rise_2023-1}.

\subsubsection{Books}
Books offer an exhaustive exploration of topics, providing context, historical perspectives, and in-depth analysis of generative AI and Fake News. 
For instance, "True Or False" by Otis (2020) guides readers in identifying Fake News, enriching the discourse with the author's intelligence analysis background\cite{otis_true_2020}. 
Similarly, "Decisively Digital" by Loth (2021) and "KI für Content Creation" by Loth (2024) delves into AI's integration in business and content creation, demonstrating these technologies' broad impact beyond Fake News detection\cite{loth_decisively_2021,loth_ki_2024}. 
Books significantly contribute to the discourse, offering nuanced insights, practical guidance, and a comprehensive understanding of generative AI and misinformation's societal implications for a broader general audience.

\section{Overview on relevant existing works}
\label{overview}

This survey methodically investigates the complex interplay between Generative Artificial Intelligence (AI) and the phenomenon of Fake News.

Structured into five interconnected sub-sections, the survey addresses:
\begin{itemize}
    \item \textbf{Enabling technologies} examines technological underpinnings such as Natural Language Processing (NLP) and Transformers, essential for Generative AI’s functionality in generating and detecting Fake News.
    \item \textbf{Creation of Fake News} discusses the dual capacity of Generative AI in content creation, focusing on its contributions to both, the generation and dissemination of content.
    \item \textbf{Case study social media as most relevant distribution channel} analyzes the influence of social media as a channel for disinformation and the effects of Fake News on public perception.
    \item \textbf{Detection of Fake News} explores methodologies and technologies for Fake News detection and analysis are, with an emphasis on Generative AI's role in augmenting these efforts.
    \item \textbf{Deepfakes as upcoming technology} discusses challenges posed by deepfakes and synthetic media, including current detection strategies and the dynamic between creation and detection.
\end{itemize}

\subsubsection{Enabling technologies}

Enabling technologies, such as Natural Language Processing (NLP) and Transformers, serve as foundations for Generative AI's evolution.
These methodologies improve the understanding and generation of human language.
Their implications for detecting, analyzing, and generating Fake News are significant.

Schütz et al. (2022) explore the efficacy of the XLM-RoBERTa model in detecting Fake News across languages, specifically targeting English and German texts\cite{schutz_ait_fhstp_2022}.
Their approach, which involves pre-training on additional datasets and fine-tuning on target data, underscores the potential of multilingual transformers in addressing the global challenge of Fake News.

Schütz et al. (2021) apply pre-trained transformer models to automatically detect Fake News, showcasing the efficiency of state-of-the-art AI technologies in discerning fact from fiction\cite{del_bimbo_automatic_2021}.

Devlin et al. (2018) introduce BERT, a transformer model that revolutionizes NLP by enabling deep bidirectional training of language representations\cite{devlin_bert_2018}.
This model lays the groundwork for sophisticated AI applications, including nuanced Fake News detection mechanisms.

In "Visual Analytics with Tableau," Loth (2019) discusses the integration of NLP techniques in data visualization through tools like Tableau, highlighting the intersection of data analysis and language processing in uncovering insights from complex datasets\cite{loth_visual_2019}.

Peters et al. (2018) provide an in-depth analysis of how contextual word embeddings, as used in models like BERT, contribute to understanding the nuances of language, a critical aspect in differentiating genuine from Fake News\cite{peters_dissecting_2018}.

Maronikolakis et al. (2021) tackle the challenge of identifying automatically generated headlines, demonstrating that transformers significantly outperform humans in distinguishing real from Fake News content\cite{maronikolakis_identifying_2021}.

De Oliveira et al. (2021) survey the application of NLP in identifying Fake News on social networks, emphasizing the importance of machine learning and language processing techniques in combating misinformation\cite{de_oliveira_identifying_2021}.

Howard and Ruder (2018) propose a method for universal language model fine-tuning, illustrating how pre-trained models can be adapted for specific tasks like text classification, including Fake News detection, enhancing the precision of AI-driven analyses\cite{howard_universal_2018}.

Vijjali et al. (2020) develop a two-stage transformer-based model for detecting COVID-19 related Fake News, combining fact-checking with textual entailment to verify claims, showcasing the adaptability of transformer models to current events and specific domains\cite{vijjali_two_2020}.

Loth (2023) examines the ethical dimensions of AI, relating technological progress to moral responsibility\cite{loth_rise_2023-1}.

Vaswani et al. (2023) propose the Transformer model\cite{vaswani_attention_2023}.
It replaces complex recurrent or convolutional networks with attention mechanisms and achieves superior performance in sequence transduction tasks.
By dispensing with recurrence and convolutions entirely, the Transformer demonstrates increased parallelizability and reduced training time.
This model not only improves the English-to-German translation task BLEU score by over 2 points but also sets a new single-model record for the English-to-French task after minimal training on eight GPUs.
The adaptability of the Transformer is further evidenced by its successful application to English constituency parsing across varying data sizes.

Bubeck et al. (2023) examine GPT-4, highlighting its near-human capabilities across multiple disciplines, indicating a move towards artificial general intelligence (AGI)\cite{bubeck_sparks_2023}.
GPT-4 emerges as a frontrunner among a new generation of LLMs, including ChatGPT and Google's PaLM, demonstrating capabilities that closely mirror human intelligence across diverse fields such as mathematics, coding, vision, medicine, law, and psychology.
GPT-4's performance, superior to predecessors, showcases its versatility in solving novel tasks without specific prompts.
This study positions GPT-4 as an early form of AGI, emphasizing its societal implications and the necessity for a new research paradigm beyond next-word prediction.

Ma et al. (2024) introduce a 1-bit Large Language Model (LLM) variant, namely BitNet b1.58, which employs ternary parameters \{-1, 0, 1\}\cite{ma_era_2024}.
This development ushers in a new era of efficiency by matching the performance of full-precision (FP16 or BF16) Transformer LLMs in both perplexity and end-task performance, despite using significantly less memory and energy.
The 1.58-bit LLM not only defines a new scaling law for training high-performance, cost-effective LLMs but also suggests a shift towards a new computational paradigm potentially supported by specialized hardware optimized for 1-bit operations.

These works collectively highlight the transformative impact of NLP and transformer models in processing and understanding complex language patterns, offering robust solutions for detecting and analyzing Fake News across various platforms and languages.

\subsubsection{Creation of Fake News}

Generative AI's role in content creation spans various domains, from aging research to synthetic data generation, highlighting its transformative impact on digital innovation and its potential ethical considerations.

Zhavoronkov et al. (2019) discuss the application of AI in aging research, demonstrating how AI can generate synthetic biological data and identify new therapeutic targets, showcasing the broader potential of generative models in biotechnology\cite{zhavoronkov_artificial_2019}.

Gozalo-Brizuela and Garrido-Merchan (2023) provide a comprehensive review of large generative AI models, such as ChatGPT and Stable Diffusion, emphasizing their revolutionary impact across various sectors and the critical need for ethical considerations\cite{gozalo-brizuela_chatgpt_2023}.

In the book "Decisively Digital," Loth (2021) explores the integration of AI in business, highlighting the strategic adoption of generative AI models like GPT-3 for enhancing digital culture and fostering innovation in content creation\cite{loth_decisively_2021}.

Jin et al. (2020) review the application of Generative Adversarial Networks (GANs) in computer vision, illustrating the versatility of GANs in image generation and style transfer, and addressing challenges such as model collapse\cite{jin_generative_2020}.

Arora and Arora (2022) highlight the role of GANs in generating synthetic patient data, underlining the benefits for clinical research and the importance of addressing ethical issues related to data privacy and authenticity\cite{arora_generative_2022}.

Radford et al. (2019) demonstrate how language models, trained on vast datasets, can perform a range of tasks without explicit supervision, suggesting a pathway for AI to learn from the diversity of human language and content\cite{radford_language_2019}.

In the book "KI für Content Creation," Loth (2024) examines the use of AI in generating diverse content types, emphasizing the technological advancements and ethical challenges posed by generative AI in creative processes\cite{loth_ki_2024}.

Weisz et al. (2023) propose design principles for generative AI applications, aiming to foster productive use while mitigating potential harms, signifying the growing need for ethical frameworks in AI development\cite{weisz_toward_2023}.

Simon et al. (2023) address fears surrounding generative AI's role in misinformation, arguing that the impact may be overestimated and calling for evidence-based approaches to understanding and mitigating potential issues\cite{simon_misinformation_2023}.

Sandrini and Somogyi (2023) analyze the effects of Generative AI advancements on news media consumption\cite{sandrini_generative_2023}.
As GenAI progresses, it initially leads consumers to increase their consumption of deceptive content, such as Fake News and clickbait.
This early-stage GenAI distorts consumer incentives and reduces welfare by encouraging the consumption of lower value goods.
However, after reaching a certain development threshold, GenAI starts to benefit consumers.
This paper recognizes that the negative effects of early-stage GenAI also include reduced investment in news production.

Ferrara (2024) highlights the potential misuse of GenAI and LLMs, where powerful capabilities bring forth significant societal challenges\cite{ferrara_genai_2024}.
Because GenAI applications can generate indistinguishable deepfakes and synthetic identities, they pose risks of enabling malicious campaigns and targeted misinformation.
This article calls attention to the urgent need to address the misuse of GenAI in creating sophisticated malware, fabricating identities, and crafting synthetic realities.
The societal implications of such misuse underscore the blurring lines between virtual and real worlds, impacting everyone.

These works collectively illuminate the dynamic interplay between generative AI and content creation, underscoring the technology's innovative potential and the critical need for responsible application and ethical consideration.

\subsubsection{Case study social media as most relevant distribution channel}

The intersection of social media, disinformation, and public perception is critical in understanding the dynamics of Fake News.
This section presents key studies that explore these themes, highlighting their contributions and findings in the broader discourse on Fake News.

Allcott and Gentzkow (2017) investigate the distribution and impact of Fake News during the 2016 U.S. presidential election.
They find that Fake News favoring Donald Trump was shared more widely on Facebook compared to content favoring Hillary Clinton.
This study underscores the significant role of social media in the dissemination of Fake News and its potential to influence public opinion and electoral outcomes\cite{allcott_social_2017}.

Bradshaw, Bailey, and Howard (2021) provide a global overview of organized social media manipulation, detailing the strategies employed by various actors to spread disinformation.
Their report highlights the industrial scale at which disinformation campaigns are orchestrated, reflecting the evolving nature of digital propaganda and its implications for democratic processes\cite{bradshaw_industrialized_2021}.

Bradshaw et al. (2022) examine the propaganda strategies of Russian-affiliated media outlets in their coverage of the \#BlackLivesMatter protests.
The study reveals divergent framing strategies among different Russian media properties, illustrating how state-backed outlets exploit social issues to influence public discourse and sow division\cite{bradshaw_playing_2022}.

Lewandowsky and Van Der Linden (2021) explore the psychological underpinnings of misinformation and propose inoculation theory as a proactive strategy to build resilience against Fake News.
Their work presents evidence that preemptive exposure to weakened forms of misinformation can "vaccinate" individuals, reducing susceptibility to Fake News\cite{lewandowsky_countering_2021}.

In the book "True Or False," Otis (2020) offers a guide to identifying Fake News, drawing on her experience as a CIA analyst.
The book provides readers with tools to discern truth from falsehood, emphasizing critical thinking and the importance of verifying sources.
Otis' work is a valuable resource for enhancing media literacy in the age of disinformation\cite{otis_true_2020}.

Shu et al. (2020) discuss various strategies for combating disinformation on social media, highlighting the roles of education, research, and collaboration. Their overview presents an array of approaches to mitigate the spread of Fake News, from technological solutions to public awareness campaigns\cite{shu_combating_2020}.

Shu et al. (2020) introduce FakeNewsNet, a comprehensive data repository containing news content, social context, and spatiotemporal information.
This resource facilitates the study of Fake News dissemination on social media, offering researchers valuable data for developing detection algorithms and understanding the dynamics of misinformation spread\cite{shu_fakenewsnet_2020}.

Yang and Tian (2021) investigate the third-person effect in the context of COVID-19 Fake News, finding that individuals perceive others as more susceptible to misinformation than themselves.
This perception may hinder efforts to address Fake News, as it leads to underestimation of one's own vulnerability to disinformation\cite{yang_others_2021}.

Pennycook and Rand (2021) delve into the psychology behind the belief and dissemination of Fake News on social media\cite{pennycook_psychology_2021}.
They challenge the prevalent view that political bias is the primary driver of susceptibility to Fake News, showing that individuals exhibit better discernment between truth and falsehood when evaluating politically aligned news.
The research highlights that poor judgment of truthfulness correlates with a lack of careful reasoning, adequate knowledge, and reliance on heuristic cues like familiarity. Moreover, they uncover a significant gap between individuals' beliefs and what they share online, attributing this to inattention rather than intentional sharing of misinformation.
This work proposes that interventions focusing on enhancing users' attention to accuracy and leveraging crowdsourced veracity ratings can improve content reliability on social media platforms.

Public perceptions of deepfake technology can significantly influence its impact on society.
For instance, \cite{verma2024one} has illustrated that even a single deceptive video could instigate severe geopolitical tensions, highlighting the urgent need for robust verification measures.

Beyond content-focused strategies, examining social signals can further enhance the detection of misinformation.
For example, \cite{alsaid2024combating} propose leveraging social noise and social entropy as measures of uncertainty, thus offering novel approaches to evaluate the credibility and trustworthiness of shared content.

These works collectively offer an understanding of the challenges posed by disinformation on social media, highlighting the importance of technological solutions, psychological resilience, and critical thinking in addressing the issue.
Each contributes unique perspectives on detection strategies, psychological defenses, and the broader societal impacts, making them particularly relevant to discussions on generative AI's role in the propagation of Fake News.

\subsubsection{Detection of Fake News}

The proliferation of Fake News and its detection have become paramount concerns in the digital age. This section reviews seminal works that have contributed significantly to the understanding and detection of Fake News.

Cao et al. (2020) delve into the significance of multimedia content in the dissemination of Fake News.
Their comprehensive review outlines the types of visual content leveraged by Fake News, effective visual features for detection, and the challenges in multimedia Fake News detection.
Unlike other works, this study emphasizes the multimedia aspect of Fake News, presenting a unique angle on detection methods\cite{cao_exploring_2020}.

Jwa et al. (2019) introduce \textit{exBAKE}, an automatic Fake News detection model leveraging Bidirectional Encoder Representations from Transformers (BERT).
By examining the relationship between the headline and body text, they achieve significant improvements in detection accuracy.
This model's reliance on deep contextual embeddings from BERT offers insights into the linguistic nuances of Fake News, setting a foundation for future research in text-based Fake News detection\cite{jwa_exbake_2019}.

Focusing on the COVID-19 pandemic, Karnyoto et al. (2022) apply BERT and GPT-2 models augmented with GRU-CRF for enhanced Fake News detection.
Their approach highlights the effectiveness of transfer learning and feature augmentation in addressing the rapid spread of misinformation during health crises\cite{karnyoto_transfer_2022}.

Lazer et al. (2018) provide a multidisciplinary analysis of Fake News, calling for collaborative efforts across fields to mitigate its impact.
Their review synthesizes research on the spread of Fake News, its societal effects, and potential detection and intervention strategies.
This work sets a broad context for understanding Fake News, emphasizing the need for a multifaceted approach to its detection\cite{lazer_science_2018}.

Liu and Wu (2018) propose a novel model utilizing recurrent and convolutional networks to classify news propagation paths for early Fake News detection on social media.
Their method underscores the importance of analyzing user engagement patterns, offering a proactive approach to curtailing the spread of misinformation\cite{liu_early_2018}.

Liu et al. (2019) present a two-stage model that leverages BERT for detecting short Fake News stories.
By treating Fake News detection as a fine-grained classification task, they address the challenge of identifying misleading information in concise texts, providing a valuable tool for real-time news verification\cite{liu_two-stage_2019}.

Novo (2021) introduces MisInfoBot, a novel system designed to combat COVID-19 misinformation on Twitter through supervised learning classification and information retrieval.
This work is particularly relevant in the context of health misinformation, offering a timely example of AI's potential to address specific misinformation challenges\cite{novo_misinfobot_2021}.

Przybyla (2020) investigates automatic methods for detecting documents of low credibility based on their writing style.
By focusing on the sensational and affective vocabulary typical of Fake News, this study contributes to understanding the stylistic markers that distinguish Fake News from credible reports\cite{przybyla_capturing_2020}.

Qian et al. (2018) leverage collective user intelligence to detect Fake News, proposing a model that generates user responses to news articles for analysis.
This innovative approach demonstrates the utility of simulating user reactions as a means to enhance Fake News detection algorithms\cite{qian_neural_2018}.

Shu et al. (2019) explore the influence of social context in detecting Fake News. Their work underscores the tri-relationship among publishers, news pieces, and users, illustrating how these dynamics can be leveraged for more accurate Fake News detection.
This approach of incorporating social media interactions provides a complementary perspective to traditional content analysis methods, highlighting the importance of considering the broader dissemination environment of Fake News\cite{shu_beyond_2019}.

Silva et al. (2020) address the challenge of Fake News in Portuguese, providing a novel dataset and a comprehensive analysis of machine learning methods for its detection.
This work not only fills a gap in language-specific resources for Fake News detection but also offers insights into the effectiveness of various classification strategies, enriching the global discourse on combating misinformation\cite{silva_towards_2020}.

Zhou and Zafarani (2019) explore the utilization of network-based clues from social media for Fake News detection, presenting a pattern-driven approach\cite{zhou_network-based_2019}.
They emphasize the patterns of Fake News propagation through the analysis of spreaders and their relationships within social networks.
This study not only adds a new dimension to Fake News detection by moving beyond content analysis but also enhances the explainability of Fake News features.
Their findings demonstrate the effectiveness of incorporating network-level patterns, including node-level, ego-level, triad-level, community-level, and overall network patterns, in improving the accuracy of Fake News detection.

While automated detection mechanisms are on the rise, human oversight remains crucial.
As \cite{goh2024humans} demonstrate, computational approaches can sometimes outperform human detection rates, yet human involvement still plays a key role in the refinement and validation of detection tools.

Each of these contributions enriches the field of Fake News detection by offering unique perspectives, methodologies, and findings.
From leveraging visual content and advanced NLP techniques to incorporating social context and user interactions, these works collectively advance our understanding of Fake News dynamics.
While each has its context and focus, together they form a multifaceted view of the challenges and opportunities in detecting Fake News, providing a solid foundation for future research in the intersection with Generative AI.

\subsubsection{Deepfakes as upcoming technology}

The advent of deepfakes and synthetic media has posed new challenges and opportunities in the domain of information security, necessitating advanced detection techniques to safeguard against the propagation of Fake News and manipulated content.

Bansal et al. (2023) explore the utilization of Convolutional Neural Networks (CNN) and Generative Adversarial Networks (DCGANs) for the detection and elimination of deepfake multimedia content, highlighting the critical role of deep learning models in mitigating the spread of fabricated information\cite{bansal_deepfake_2023}.

Botha and Pieterse (2020) provide an overview of the creation and detection methods for Fake News and deepfakes, emphasizing the significant threat they pose to 21st-century information security and the necessity for robust detection tools to distinguish between real and synthetic content\cite{botha_fake_2020}.

Wang et al. (2020) propose FakeSpotter, a novel approach based on monitoring neuron behaviors, to detect AI-synthesized fake faces, demonstrating that deep learning systems can effectively identify manipulations in facial images, thus contributing to the integrity of visual information\cite{wang_fakespotter_2020}.

Cardenuto et al. (2023) delve into the implications of synthetic realities generated through AI, underlining the urgency for digital forensic techniques that can discern harmful synthetic creations from reality, especially in the context of combating Fake News and misinformation\cite{cardenuto_age_2023}.

Mosallanezhad et al. (2020) tackle the generation of topic-preserving synthetic news via an adversarial deep reinforcement learning approach, illustrating the dual-use potential of AI in both generating realistic news content and developing methodologies for its detection\cite{mosallanezhad_topic-preserving_2020}.

Cocchi et al. (2023) conducte an experimental analysis on the robustness of deepfake detection methods against various image transformations, revealing the susceptibility of detection models to manipulation and the need for adaptable and resilient detection algorithms\cite{cocchi_unveiling_2023}.

As shown by \cite{chun2024can}, older adults may face unique challenges in spotting deepfakes due to their differing identification strategies and experience with digital content.
Moreover, demographic differences in detection abilities, such as those highlighted by \cite{chun2024can}, underscore the need for tailored training and resources for various user groups.

These works collectively underscore the escalating arms race between the creation of deepfakes and the development of detection technologies.
They highlight the challenges presented by synthetic media and the imperative for continuous innovation in detection methodologies to ensure the veracity of digital content in the age of generative AI.

\section{Findings}
\label{findings}

\subsection{Evolution of the topic}
\label{survey:evolution}



The landscape of research surrounding Generative AI and its implications for Fake News has undergone significant transformation from 2022 to 2024.
Initially, the focus was predominantly on understanding the basics of Generative AI technologies and their potential applications.
Early research concentrated on foundational models and techniques, such as those discussed in \cite{radford_language_2019} and \cite{devlin_bert_2018}, which laid the groundwork for later advancements in content generation and analysis.

As the technology matured, attention shifted towards exploiting Generative AI for more specific tasks. The creation and detection of synthetic media, including deepfakes, emerged as a critical area of investigation, with significant contributions from \cite{bansal_deepfake_2023} and \cite{botha_fake_2020}. Parallel to these technical explorations, researchers also delved into the societal impact of Fake News, scrutinizing its influence on public perception and social media dynamics, as highlighted by \cite{allcott_social_2017} and \cite{bradshaw_industrialized_2021}.

More recently, the focus has expanded to encompass ethical considerations and mitigation strategies, addressing the broader consequences of Generative AI's capabilities. Studies such as \cite{simon_misinformation_2023} and \cite{weisz_toward_2023} underscore the importance of responsible AI use and the development of frameworks to counter the proliferation of Fake News effectively.

The evolution of this research topic reflects in the diversification of domains and methodologies, as indicated in Table~\ref{tab:topics}. Initially centered on the technical aspects of Generative AI, the research community gradually embraced a more holistic approach. This shift acknowledges the multifaceted nature of Fake News, incorporating technological, social, and ethical dimensions into the discourse.

The landscape of Generative AI in the context of Fake News encompasses significant advancements and challenges in March 2024.
Sophisticated Generative AI models, such as the latest GPT iterations, advanced autoencoders, and the innovative 1-bit LLMs introduced by Ma et al. (2024)\cite{ma_era_2024}, now produce hyper-realistic Fake News content in text, image, audio, and video formats\cite{ferrara_genai_2024, loth_ki_2024}.

\begin{table*}[t]
\centering
\resizebox{\textwidth}{!}{
\begin{tabular}{|p{2.8cm}|p{1.8cm}|p{1.8cm}|p{1.8cm}|p{1.8cm}|p{1.8cm}|p{1.8cm}|p{1.8cm}|p{1.8cm}|}
\toprule
Year & 2017 & 2018 & 2019 & 2020 & 2021 & 2022 & 2023 & 2024 \\
\midrule
Enabling technologies &
&
\cite{devlin_bert_2018}, \cite{howard_universal_2018}, \cite{peters_dissecting_2018} &
\cite{leo_machine_2019}, \cite{loth_visual_2019} &
\cite{oshikawa_survey_2020}, \cite{vijjali_two_2020} &
\cite{de_oliveira_identifying_2021}, \cite{kalyan_ammus_2021}, \cite{maronikolakis_identifying_2021}, \cite{del_bimbo_automatic_2021} &
\cite{schutz_ait_fhstp_2022} &
\cite{loth_rise_2023-1}, \cite{vaswani_attention_2023}, \cite{bubeck_sparks_2023} &
\cite{ma_era_2024}
 \\
  \hline
Creation of Fake News &
&
&
\cite{radford_language_2019}, \cite{zhavoronkov_artificial_2019} &
\cite{jin_generative_2020} &
\cite{loth_decisively_2021} &
\cite{arora_generative_2022} &
\cite{cao_comprehensive_2023}, \cite{gozalo-brizuela_chatgpt_2023}, \cite{simon_misinformation_2023}, \cite{weisz_toward_2023}, \cite{sandrini_generative_2023} &
\cite{loth_ki_2024}, \cite{ferrara_genai_2024}
\\
 \hline
Case study social media as most relevant distribution channel & \cite{allcott_social_2017} &
 &
 &
\cite{otis_true_2020}, \cite{shu_combating_2020}, \cite{shu_fakenewsnet_2020}, \cite{pennycook_who_2020} &
\cite{bradshaw_industrialized_2021}, \cite{lewandowsky_countering_2021}, \cite{yang_others_2021} &
\cite{bradshaw_playing_2022} &
\cite{melchior_systematic_2023} &
\cite{verma2024one}, \cite{alsaid2024combating}
 \\
 \hline
Detection of Fake News & \cite{shu_fake_2017} &
\cite{lazer_science_2018}, \cite{liu_early_2018}, \cite{qian_neural_2018} &
\cite{bondielli_survey_2019}, \cite{jwa_exbake_2019}, \cite{liu_two-stage_2019}, \cite{sharma_combating_2019}, \cite{shu_beyond_2019}, \cite{zhou_network-based_2019} &
\cite{cao_exploring_2020}, \cite{przybyla_capturing_2020}, \cite{silva_towards_2020}, \cite{zhou_survey_2020} &
\cite{novo_misinfobot_2021} &
\cite{karnyoto_transfer_2022} &
\cite{villela_fake_2023} &
\cite{goh2024humans}
 \\
 \hline
Deepfakes as upcoming technology &
 &
 &
 &
\cite{botha_fake_2020}, \cite{mosallanezhad_topic-preserving_2020}, \cite{wang_fakespotter_2020} &
 &
 &
\cite{bansal_deepfake_2023}, \cite{cardenuto_age_2023}, \cite{cocchi_unveiling_2023}, \cite{khanjani_audio_2023} &
\cite{chun2024can}
 \\
\bottomrule
\end{tabular}
}
\caption{Domains over time.}
\label{tab:topics}
\end{table*}

\subsection{Relevant venues}
\label{survey:venues}





The landscape of research pertaining to Generative AI and Fake News detection diverts across a variety of scholarly venues.
This reflects the novelty and resulting fragmentation of this multidisciplinary field and underscores the importance nature of the topic.
Table~\ref{tab:venue_citations} presents a summary of the venues with the associated citations, offering a panoramic view of where significant research contributions have been presented.

From the data collated, it is evident that the \textit{Proceedings of the Association for Information Science and Technology}, \textit{Association for Computational Linguistics} and the \textit{AAAI Conference on Artificial Intelligence} are prominent platforms for presenting seminal works, indicative of the strong reliance on advanced computational methods in this domain. The recurring presence of certain conferences, such as the \textit{International Conference on Pattern Recognition} and the \textit{International Joint Conference on Artificial Intelligence}, highlights the importance of AI and machine learning techniques in the development and assessment of Fake News detection methodologies.

Furthermore, the \textit{Trends in Cognitive Sciences} and \textit{Information Sciences} journals stand out as critical repositories of research that offer theoretical frameworks and empirical studies that advance our understanding of Fake News dynamics and its detection. These venues, along with the \textit{Science} journal, have been instrumental in disseminating research that bridges the gap between technical solutions and their psychological and societal implications.

Clusters can be observed within certain conference series, suggesting a community of researchers consistently contributing to and building upon the collective knowledge in the field. For instance, the consistent contributions to the \textit{ACM International Conference on Multimedia} and various ACM-sponsored outlets reflect a focus on multimedia aspects in Fake News, such as deepfake detection. 

The diversity of venues also reflects the broad spectrum of Generative AI applications, from content creation in \textit{Harvard Kennedy School (HKS) Misinformation Review} to ethical considerations in the \textit{Workshop on Human-AI Co-Creation with Generative Models (HAI-GEN)}. These patterns reveal the depth and complexity of the research, demanding a cross-sectional approach that spans technical, ethical, and regulatory dimensions.

\begin{table*}[t]
\centering
\begin{tabularx}{\textwidth}{|c|X|c|}
\hline
\textbf{Type} & \textbf{Venue} & \textbf{References} \\ 
\hline
Journal & Proceedings of the Association for Information Science and Technology & \cite{chun2024can}, \cite{verma2024one}, \cite{goh2024humans}, \cite{alsaid2024combating}\\
\hline
Conference & Association for Computational Linguistics (ACL) & \cite{howard_universal_2018}, \cite{oshikawa_survey_2020}, \cite{peters_dissecting_2018} \\
\hline
Conference & AAAI Conference on Artificial Intelligence & \cite{liu_early_2018}, \cite{przybyla_capturing_2020} \\
\hline
Journal & Trends in Cognitive Sciences & \cite{bradshaw_industrialized_2021}, \cite{pennycook_who_2020} \\
\hline
Conference & International Joint Conference on Artificial Intelligence & \cite{qian_neural_2018}, \cite{wang_fakespotter_2020} \\
\hline
Conference & International Conference on Pattern Recognition & \cite{del_bimbo_automatic_2021}, \cite{schutz_ait_fhstp_2022} \\
\hline
Journal & Information Sciences & \cite{bondielli_survey_2019} \\
\hline
Journal & The International Journal of Press/Politics & \cite{bradshaw_playing_2022} \\
\hline
Journal & Science & \cite{lazer_science_2018} \\
\hline
Journal & Sage New Media \& Society & \cite{melchior_systematic_2023} \\
\hline
Journal & MDPI Risks & \cite{leo_machine_2019} \\
\hline
Journal & MDPI Information & \cite{de_oliveira_identifying_2021} \\
\hline
Journal & MDPI Applied Sciences & \cite{jwa_exbake_2019} \\
\hline
Journal & Journal on Interactive Systems & \cite{villela_fake_2023} \\
\hline
Conference & International Conference on Knowledge Science, Engineering and Management & \cite{liu_two-stage_2019} \\
\hline
Conference & International Conference on IoT, Communication and Automation Technology (ICICAT) & \cite{bansal_deepfake_2023} \\
\hline
Conference & International Conference on Cyber Warfare and Security & \cite{botha_fake_2020} \\
\hline
Conference & Image Analysis and Processing – ICIAP & \cite{cocchi_unveiling_2023} \\
\hline
Journal & ACM Computing Surveys & \cite{zhou_survey_2020} \\
\hline
Journal & Harvard Kennedy School (HKS) Misinformation Review & \cite{simon_misinformation_2023} \\
\hline
Journal & Future Healthcare Journal & \cite{arora_generative_2022} \\
\hline
Journal & Frontiers in Big Data & \cite{khanjani_audio_2023} \\
\hline
Journal & Expert Systems with Applications & \cite{silva_towards_2020} \\
\hline
Journal & European Review of Social Psychology & \cite{lewandowsky_countering_2021} \\
\hline
Journal & Computers in Human Behavior & \cite{yang_others_2021} \\
\hline
Journal & Computer Science and Information Systems (ComSIS) & \cite{karnyoto_transfer_2022} \\
\hline
Journal & Ageing Research Reviews (ARR) & \cite{zhavoronkov_artificial_2019} \\
\hline
Journal & ACM Transactions on Intelligent Systems and Technology & \cite{sharma_combating_2019} \\
\hline
Journal & ACM SIGKDD Explorations & \cite{shu_fake_2017} \\
\hline
Conference & ACM International Conference on Web Search and Data Mining & \cite{shu_beyond_2019} \\
\hline
Conference & ACM International Conference on Multimedia & \cite{jin_generative_2020} \\
\hline
Workshop & Workshop on Human-AI Co-Creation with Generative Models (HAI-GEN) & \cite{weisz_toward_2023} \\
\hline
\end{tabularx}
\caption{Relevant venues.}
\label{tab:venue_citations}
\end{table*}

\subsection{Most active groups}
\label{survey:groups}



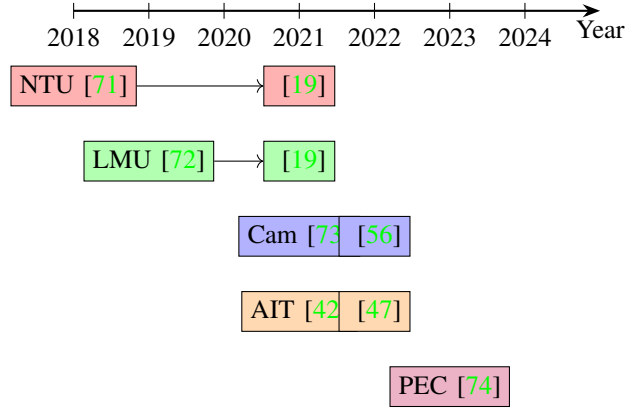
\begin{figure}[htbp]
\centering
\begin{tikzpicture}
    \draw[thick, -Stealth] (0,0) -- (7,0) node[anchor=north] {Year};

    \foreach \x/\year in {0/2018, 1/2019, 2/2020, 3/2021, 4/2022, 5/2023, 6/2024} {
        \draw (\x cm,3pt) -- (\x cm,-3pt) node[anchor=north] {\year};
    }

    \node[draw, rectangle, fill=red!30] (NTU1) at (0,-1) {NTU \cite{tandoc_jr_defining_2018}};
    \node[draw, rectangle, fill=red!30] (NTU2) at (3,-1) {\cite{maronikolakis_identifying_2021}};

    \node[draw, rectangle, fill=green!30] (LMU1) at (1,-2) {LMU \cite{egelhofer_fake_2019}};
    \node[draw, rectangle, fill=green!30] (LMU2) at (3,-2) {\cite{maronikolakis_identifying_2021}};

    \node[draw, rectangle, fill=blue!30] (Cam1) at (3,-3) {Cam \cite{osmundsen_partisan_2021}};
    \node[draw, rectangle, fill=blue!30] (Cam2) at (4.3,-3) {\cite{arora_generative_2022}};

    \node[draw, rectangle, fill=orange!30] (AIT1) at (3,-4) {AIT \cite{del_bimbo_automatic_2021}};
    \node[draw, rectangle, fill=orange!30] (AIT2) at (4.3,-4) {\cite{schutz_ait_fhstp_2022}};

    \node[draw, rectangle, fill=purple!30] (PEC1) at (5,-5) {PEC \cite{noauthor_podexcellence_2023}};
    \node[draw, rectangle, fill=purple!30] (PEC2) at (6.3,-5) {\cite{noauthor_podexcellence_2024}};

    \draw[->] (NTU1) -- (NTU2);
    \draw[->] (LMU1) -- (LMU2);
\end{tikzpicture}
\caption{Timeline of research activities by leading groups from 2018 to 2024, illustrating the dynamic landscape of contributions in the fields of digital forensics, disinformation detection, and generative AI. Each colored box represents a significant milestone or publication by the respective research group.}
\label{fig:research-activity-timeline}
\end{figure}

The research landscape in generative AI, deepfakes, and Fake News detection features contributions from a select number of academic groups and institutions globally.
These entities have collectively propelled advancements in understanding and mitigating misinformation's effects.
Figure \ref{fig:research-activity-timeline} illustrates the evolution and focus of research activities by leading groups from 2018 to 2024, underlining the growing efforts to combat disinformation and enhance generative AI technologies.
Aside from this survey, only a few groups delve into generative AI within the context of Fake News, indicating the field's emerging nature.
The involvement of multiple groups, rather than isolated researchers, adds significant weight to the research conducted in this domain.

Research on generative AI's role in misinformation involves notable academic and technological institutions, divided into universities and other research entities.

\subsubsection{Universities}

\paragraph{Cambridge University}
Cambridge University's research team explores generative AI's societal impacts, focusing on misinformation. 
They investigate both theoretical aspects and AI's practical uses against Fake News, with key studies addressing psychological motivations for sharing political Fake News \cite{osmundsen_partisan_2021} and utilizing generative adversarial networks for synthetic patient data generation \cite{arora_generative_2022}.

\paragraph{Nanyang Technological University, Singapore}
At Nanyang Technological University, Singapore, researchers contribute significantly to AI and media integrity. 
Their innovations include FakeSpotter for identifying AI-generated fake images \cite{wang_fakespotter_2020} and studies on Fake News definitions \cite{tandoc_jr_defining_2018}.

\paragraph{Ludwig-Maximilians-University of Munich}
This group uses NLP and machine learning for misinformation detection, notably in identifying AI-generated content \cite{maronikolakis_identifying_2021} and analyzing Fake News discourse \cite{egelhofer_fake_2019}.

\subsubsection{Other Research Entities}

\paragraph{Austrian Institute of Technology}
Specializing in cross-lingual Fake News detection, the Austrian Institute of Technology employs transformers to identify misinformation in multiple languages \cite{schutz_ait_fhstp_2022} and explores automatic Fake News detection \cite{del_bimbo_automatic_2021}.

\paragraph{Pôle d'excellence cyber (PEC)}
Focusing on cybersecurity and information manipulation, Pôle d'excellence cyber (PEC) provides insights into combating information manipulation through comprehensive expert reports. 
Their contributions, while not exclusively on generative AI, are vital for understanding cybersecurity aspects of Fake News and deepfakes\cite{noauthor_podexcellence_2023,noauthor_podexcellence_2024}.

These groups contribute to a dynamic ecosystem focused on combating misinformation, a critical issue in the digital era.
Their efforts underscore the interdisciplinary approach required to tackle Fake News, merging insights from computer science, psychology, media studies, and cybersecurity.

\subsection{Open Issues}
\label{survey:open}


The emergence of Generative AI has opened new frontiers in content creation, including the generation of human-like text, images, and videos\cite{loth_ki_2024}.
The intersection of Generative AI and Fake News is a field, marked by several unresolved issues that underscore the complexity of the benefits and risks of advanced AI technologies.

One pressing unresolved issue is the specific application of Generative AI in Fake News creation, which, while showcasing technological advancement, also poses significant risks to information integrity\cite{cao_exploring_2020,jwa_exbake_2019}.
While they offer significant potential for positive applications, their capacity for creating convincingly realistic fake content poses severe threats to informational integrity and public trust \cite{gozalo-brizuela_chatgpt_2023}.
The technological arms race between Generative AI advancements and detection countermeasures underscores a significant research gap\cite{karnyoto_transfer_2022,lazer_science_2018}.
Challenges include the rapid evolution of Generative AI, ethical concerns, and the scarcity of comprehensive datasets\cite{liu_early_2018,liu_two-stage_2019}.

\subsubsection{Notable Gap in Literature: Generative AI and Fake News}
The literature has yet to fully explore the intersection of Generative AI and Fake News. This gap is attributed to the nascent stage of these technologies and the complexity of predicting their misuse \cite{weisz_toward_2023,simon_misinformation_2023}.

Future studies should focus on simulative research, cross-domain analysis, and the exploration of both technological and social countermeasures. This comprehensive approach is essential for safeguarding information integrity and maintaining public trust in the age of advanced AI \cite{cardenuto_age_2023,mosallanezhad_topic-preserving_2020}.

Addressing the synergy between Generative AI technologies and the propagation of Fake News is critical.
Future research must navigate not only technological innovations but also the broader societal, ethical, and psychological dimensions of this challenge.


\subsection{The Next Frontiers in Fake News and Generative AI}

Researchers initially focused on foundational Generative AI technologies. 
\cite{radford_language_2019} and \cite{devlin_bert_2018} provided the early groundwork. 
As Generative AI matured, the research became more and more focused on its application in creating and detecting synthetic media. 
Significant contributions in this phase included \cite{bansal_deepfake_2023} and \cite{botha_fake_2020}. 
Concurrently, the societal impact of Fake News received attention, with studies like \cite{allcott_social_2017} and \cite{bradshaw_industrialized_2021} examining its effects on public perception and social media. 

Recent research has broadened to include ethical considerations and mitigation strategies. 
Works such as \cite{simon_misinformation_2023} and \cite{weisz_toward_2023} emphasized responsible AI use. 
This topic's evolution is marked by a diversification in domains and methodologies. 
The community transitioned from a technical to a holistic approach, recognizing the complex nature of Fake News. 
By March 2024, the landscape featured advanced Generative AI models capable of producing hyper-realistic content across multiple formats \cite{ferrara_genai_2024, loth_ki_2024}.
It highlights the increasing sophistication of Generative AI applications in the realm of Fake News, pointing to both the technological prowess and the attendant ethical, societal, and technical challenges.

\section{State of the Research}
\label{state}

\subsection{Impact of Generative AI on news}

Generative AI significantly impacts news production, from creating innovative content to spreading misinformation.

Models like BERT have pushed the boundaries of natural language understanding and generation, simulating human-like articles\cite{devlin_bert_2018}.
Gozalo-Brizuela and Garrido-Merchan (2023) highlight Generative AI's role in diverse content creation across sectors\cite{gozalo-brizuela_chatgpt_2023}.

Bansal et al. (2023) and Wang et al. (2020) demonstrate Generative AI's ability to produce realistic synthetic media, raising concerns about deepfakes\cite{bansal_deepfake_2023,wang_fakespotter_2020}.
This advancement challenges the integrity of information dissemination, necessitating robust Fake News detection methods\cite{botha_fake_2020,cardenuto_age_2023}.

Mosallanezhad et al. (2020) show Generative AI's sophistication in creating contextually relevant synthetic news, blurring the lines between real and AI-generated content\cite{mosallanezhad_topic-preserving_2020}.

Simon et al. (2023) discuss the ethical considerations of Generative AI in news, emphasizing the need for balance between its benefits and risks\cite{simon_misinformation_2023}.
The dual-use nature of Generative AI poses ethical dilemmas, requiring careful application in journalism to maintain trust and credibility.

\subsection{Mitigation strategies}

Generative AI's innovative potential for content creation also necessitates robust mitigation strategies against Fake News and misinformation.

Lewandowsky and Van Der Linden (2021) propose inoculation theory, a psychological approach to build resilience against misinformation\cite{lewandowsky_countering_2021}.
This approach emphasizes preemptive education and awareness to mitigate Fake News impacts.

Weisz et al. (2023) suggest design principles for generative AI that prioritize safety and ethical use\cite{weisz_toward_2023}.
These principles focus on preventing harmful outputs and misuse in AI systems.

Shu et al. (2020) explore collaborative measures against disinformation on social media, involving governments, private sector, and civil society\cite{shu_combating_2020}.
Collaboration is key to forming an effective response against misinformation.

Simon et al. (2023) highlight the need for evidence-based methods to address fears about generative AI's misinformation role\cite{simon_misinformation_2023}.
Dialogue among academics, policymakers, and developers is crucial for addressing misinformation complexities.

Advanced detection algorithms, like those developed by Jwa et al. (2019) and Bansal et al. (2023), are vital for identifying AI-generated Fake News\cite{jwa_exbake_2019,bansal_deepfake_2023}.
The progress of these technologies is essential for maintaining digital information integrity.

Countering Generative AI's adverse effects on news requires diverse strategies, including technology, education, ethics, and collaboration.
As Generative AI evolves, adaptation and innovation in mitigation strategies are imperative.

\subsection{Risks and stakeholders}

Generative AI revolutionizes content creation and introduces risks requiring stakeholder reevaluation.
This synthesis examines risks from Generative AI in news and identifies stakeholders involved in mitigation.
Generative AI generates realistic Fake News and deepfakes, threatening information integrity and public trust.
Bradshaw et al. (2021) show social media manipulation campaigns using Generative AI to spread global disinformation\cite{bradshaw_industrialized_2021}.
Botha and Pieterse (2020) emphasize the threat of Fake News to information security, highlighting detection technology's role\cite{botha_fake_2020}.

Stakeholders include academics, technology developers, social media platforms, policymakers, and the public.
Academics develop detection algorithms and study misinformation's psychological aspects, proposing strategies like inoculation theory\cite{lewandowsky_countering_2021}.

Technology developers and platforms play crucial roles in deploying AI detection systems and enforcing misuse prevention policies.
Weisz et al. (2023) suggest design principles for ethical AI application, guiding responsible development\cite{weisz_toward_2023}.

Policymakers create regulations that protect public interests while fostering innovation in AI content generation.
Otis (2020) advocates for public awareness and media literacy to help individuals discern information authenticity\cite{otis_true_2020}.

Collaboration among stakeholders is vital for a comprehensive approach to Generative AI risks in news.
Fostering dialogue, research, and policy development, stakeholders aim to minimize AI-generated misinformation impacts while leveraging technology's benefits.

\section{Conclusion}
\label{conclusion}

This survey has undertaken a systematic exploration of the interplay between Generative AI and Fake News, filling a critical void in the existing corpus of literature.
In \Cref{related}, this survey has substantiated our methodological approach, highlighting the essential nature of this exhaustive survey within the dynamic environment of Generative AI and misinformation.
By categorizing the existing body of work, this survey revealed diverse methodologies in the fight against Fake News, facilitated by Generative AI.
\Cref{domain} established a foundational comprehension of the domain, shedding light on pivotal concepts and aspects crucial for understanding the broader ramifications of Generative AI in the realm of Fake News.
It presented a comprehensive structural analysis of the domain, identifying gaps that were previously unaddressed.
\Cref{methodology} delineated the methodological framework employed in this investigation, guaranteeing a methodical exploration.

\Cref{overview} delivered a comprehensive review of recent scholarly endeavors, accentuating the progress achieved in leveraging Generative AI for both the detection and creation of Fake News.
A systematic enumeration of leading venues and research groups showcased the focal points of current research, highlighting where significant advancements were being made.
\Cref{findings} charted the developmental trajectory of this research area, spotlighting noteworthy venues and the contributions of eminent research collectives.
The survey brought to light unresolved issues within the domain, suggesting avenues for future research and setting a clear direction for scholarly inquiry.
It also spotlighted pressing research queries that propel further scholarly endeavors.
\Cref{state} amalgamated the insights derived from the survey, presenting a condensed perspective of the contemporary state of the art and delineating prospective avenues for domain advancement.
By pinpointing the state-of-the-art, it outlined the forefront of technological advancements in Generative AI, emphasizing its dual role in both generating and detecting Fake News.
These contributions collectively enhanced our understanding of the domain, providing a valuable resource for researchers, practitioners, and policy makers engaged in the ongoing battle against misinformation.


Section~\ref{findings} charted the trajectory of research on Generative AI and its interplay with Fake News.
Over time, the focus of Generative AI research expanded from foundational model building to applying these models for both creation and detection of synthetic media.
This gradual shift led to a broader awareness of the ethical, societal, and mitigation dimensions of Fake News, reflecting a transition from purely technical solutions to a more comprehensive perspective on hyper-realistic AI-generated content and its public impact.
By early 2024, advanced models capable of producing highly realistic content across multiple formats highlighted the evolving sophistication of Generative AI systems.
This progression underscored both the technological advancements achieved and the imperative for responsible governance to address associated challenges.


This survey stands as a reference within the domain of Generative AI and Fake News, offering insights and directions for tackling misinformation.
Its comprehensive analysis and categorization illuminated the interplay between AI technologies and the dissemination of Fake News.
By highlighting active research groups and prominent venues, the survey fostered a collaborative research environment poised for breakthroughs in misinformation detection and mitigation.

The identification of unresolved issues and future research directions will serve to mature and propel the field forward.
It emphasizes the critical need for innovative solutions for handling Fake News which is a key challenge for our society at this generation.

\bibliographystyle{IEEEtran}
\bibliography{bibliography}

\begin{thebibliography}{10}
\providecommand{\url}[1]{#1}
\csname url@samestyle\endcsname
\providecommand{\newblock}{\relax}
\providecommand{\bibinfo}[2]{#2}
\providecommand{\BIBentrySTDinterwordspacing}{\spaceskip=0pt\relax}
\providecommand{\BIBentryALTinterwordstretchfactor}{4}
\providecommand{\BIBentryALTinterwordspacing}{\spaceskip=\fontdimen2\font plus
\BIBentryALTinterwordstretchfactor\fontdimen3\font minus \fontdimen4\font\relax}
\providecommand{\BIBforeignlanguage}[2]{{%
\expandafter\ifx\csname l@#1\endcsname\relax
\typeout{** WARNING: IEEEtran.bst: No hyphenation pattern has been}%
\typeout{** loaded for the language `#1'. Using the pattern for}%
\typeout{** the default language instead.}%
\else
\language=\csname l@#1\endcsname
\fi
#2}}
\providecommand{\BIBdecl}{\relax}
\BIBdecl

\bibitem{berghel_malice_2018}
\BIBentryALTinterwordspacing
H.~Berghel, ``Malice domestic: The cambridge analytica dystopia,'' vol.~51, no.~5, pp. 84--89, conference Name: Computer. [Online]. Available: \url{https://ieeexplore.ieee.org/document/8364652}
\BIBentrySTDinterwordspacing

\bibitem{loth_openai_2023}
\BIBentryALTinterwordspacing
A.~Loth. {OpenAI} {GPT}-4: Exploring the evolution and impact of generative {AI}. [Online]. Available: \url{https://alexloth.com/gpt-4-launches-today-the-rise-of-generative-ai-from-neural-networks-to-deepmind-and-openai/}
\BIBentrySTDinterwordspacing

\bibitem{shu_fake_2017}
\BIBentryALTinterwordspacing
K.~Shu, A.~Sliva, S.~Wang, J.~Tang, and H.~Liu, ``Fake news detection on social media: A data mining perspective.'' [Online]. Available: \url{http://arxiv.org/abs/1708.01967}
\BIBentrySTDinterwordspacing

\bibitem{wardle2017information}
C.~Wardle and H.~Derakhshan, \emph{Information disorder: Toward an interdisciplinary framework for research and policymaking}.\hskip 1em plus 0.5em minus 0.4em\relax Council of Europe Strasbourg, 2017, vol.~27.

\bibitem{sharma_combating_2019}
\BIBentryALTinterwordspacing
K.~Sharma, F.~Qian, H.~Jiang, N.~Ruchansky, M.~Zhang, and Y.~Liu, ``Combating fake news: A survey on identification and mitigation techniques,'' vol.~10, no.~3, pp. 21:1--21:42. [Online]. Available: \url{https://doi.org/10.1145/3305260}
\BIBentrySTDinterwordspacing

\bibitem{oshikawa_survey_2020}
\BIBentryALTinterwordspacing
R.~Oshikawa, J.~Qian, and W.~Y. Wang, ``A survey on natural language processing for fake news detection.'' [Online]. Available: \url{http://arxiv.org/abs/1811.00770}
\BIBentrySTDinterwordspacing

\bibitem{villela_fake_2023}
\BIBentryALTinterwordspacing
H.~F. Villela, F.~Corrêa, J.~S. d. A.~N. Ribeiro, A.~Rabelo, and D.~B.~F. Carvalho, ``Fake news detection: a systematic literature review of machine learning algorithms and datasets,'' vol.~14, no.~1, pp. 47--58, number: 1. [Online]. Available: \url{https://sol.sbc.org.br/journals/index.php/jis/article/view/3020}
\BIBentrySTDinterwordspacing

\bibitem{leo_machine_2019}
M.~Leo, S.~Sharma, and K.~Maddulety, ``Machine learning in banking risk management: A literature review,'' vol.~7, no.~1, p.~29, publisher: {MDPI}.

\bibitem{cao_comprehensive_2023}
\BIBentryALTinterwordspacing
Y.~Cao, S.~Li, Y.~Liu, Z.~Yan, Y.~Dai, P.~S. Yu, and L.~Sun, ``A comprehensive survey of {AI}-generated content ({AIGC}): A history of generative {AI} from {GAN} to {ChatGPT}.'' [Online]. Available: \url{http://arxiv.org/abs/2303.04226}
\BIBentrySTDinterwordspacing

\bibitem{kalyan_ammus_2021}
K.~S. Kalyan, A.~Rajasekharan, and S.~Sangeetha, ``Ammus: A survey of transformer-based pretrained models in natural language processing.''

\bibitem{melchior_systematic_2023}
\BIBentryALTinterwordspacing
C.~Melchior and M.~Oliveira, ``A systematic literature review of the motivations to share fake news on social media platforms and how to fight them,'' p. 14614448231174224, publisher: {SAGE} Publications. [Online]. Available: \url{https://doi.org/10.1177/14614448231174224}
\BIBentrySTDinterwordspacing

\bibitem{zhou_survey_2020}
\BIBentryALTinterwordspacing
X.~Zhou and R.~Zafarani, ``A survey of fake news: Fundamental theories, detection methods, and opportunities,'' vol.~53, no.~5, pp. 109:1--109:40. [Online]. Available: \url{https://doi.org/10.1145/3395046}
\BIBentrySTDinterwordspacing

\bibitem{bondielli_survey_2019}
\BIBentryALTinterwordspacing
A.~Bondielli and F.~Marcelloni, ``A survey on fake news and rumour detection techniques,'' vol. 497, pp. 38--55. [Online]. Available: \url{https://www.sciencedirect.com/science/article/pii/S0020025519304372}
\BIBentrySTDinterwordspacing

\bibitem{khanjani_audio_2023}
\BIBentryALTinterwordspacing
Z.~Khanjani, G.~Watson, and V.~P. Janeja, ``Audio deepfakes: A survey,'' vol.~5, p. 1001063. [Online]. Available: \url{https://www.ncbi.nlm.nih.gov/pmc/articles/PMC9869423/}
\BIBentrySTDinterwordspacing

\bibitem{biever_chatgpt_2023}
\BIBentryALTinterwordspacing
C.~Biever, ``{ChatGPT} broke the turing test — the race is on for new ways to assess {AI},'' vol. 619, no. 7971, pp. 686--689, bandiera\_abtest: a Cg\_type: News Feature Number: 7971 Publisher: Nature Publishing Group Subject\_term: Computer science, Mathematics and computing, Technology, Society. [Online]. Available: \url{https://www.nature.com/articles/d41586-023-02361-7}
\BIBentrySTDinterwordspacing

\bibitem{gozalo-brizuela_chatgpt_2023}
\BIBentryALTinterwordspacing
R.~Gozalo-Brizuela and E.~C. Garrido-Merchan, ``{ChatGPT} is not all you need. a state of the art review of large generative {AI} models.'' [Online]. Available: \url{http://arxiv.org/abs/2301.04655}
\BIBentrySTDinterwordspacing

\bibitem{sandner_blockchain_2024}
\BIBentryALTinterwordspacing
P.~Sandner and A.~Loth, ``The future of blockchain: A decisively digital conversation with philipp sandner,'' January 2024. [Online]. Available: \url{https://alexloth.com/the-future-of-blockchain-a-decisively-digital-conversation-with-philipp-sandner/}
\BIBentrySTDinterwordspacing

\bibitem{devlin_bert_2018}
J.~Devlin, M.-W. Chang, K.~Lee, and K.~Toutanova, ``Bert: Pre-training of deep bidirectional transformers for language understanding.''

\bibitem{jwa_exbake_2019}
\BIBentryALTinterwordspacing
H.~Jwa, D.~Oh, K.~Park, J.~M. Kang, and H.~Lim, ``{exBAKE}: Automatic fake news detection model based on bidirectional encoder representations from transformers ({BERT}),'' vol.~9, no.~19, p. 4062, number: 19 Publisher: Multidisciplinary Digital Publishing Institute. [Online]. Available: \url{https://www.mdpi.com/2076-3417/9/19/4062}
\BIBentrySTDinterwordspacing

\bibitem{maronikolakis_identifying_2021}
\BIBentryALTinterwordspacing
A.~Maronikolakis, H.~Schutze, and M.~Stevenson, ``Identifying automatically generated headlines using transformers.'' [Online]. Available: \url{http://arxiv.org/abs/2009.13375}
\BIBentrySTDinterwordspacing

\bibitem{cao_exploring_2020}
\BIBentryALTinterwordspacing
J.~Cao, P.~Qi, Q.~Sheng, T.~Yang, J.~Guo, and J.~Li, ``Exploring the role of visual content in fake news detection.'' [Online]. Available: \url{http://arxiv.org/abs/2003.05096}
\BIBentrySTDinterwordspacing

\bibitem{liu_two-stage_2019}
C.~Liu, X.~Wu, M.~Yu, G.~Li, J.~Jiang, W.~Huang, and X.~Lu, ``A two-stage model based on {BERT} for short fake news detection,'' in \emph{Knowledge Science, Engineering and Management}, ser. Lecture Notes in Computer Science, C.~Douligeris, D.~Karagiannis, and D.~Apostolou, Eds.\hskip 1em plus 0.5em minus 0.4em\relax Springer International Publishing, pp. 172--183.

\bibitem{karnyoto_transfer_2022}
\BIBentryALTinterwordspacing
A.~Karnyoto, C.~Sun, B.~Liu, and X.~Wang, ``Transfer learning and {GRU}-{CRF} augmentation for covid-19 fake news detection,'' vol.~19, no.~2, pp. 639--658. [Online]. Available: \url{https://doiserbia.nb.rs/Article.aspx?ID=1820-02142100053K}
\BIBentrySTDinterwordspacing

\bibitem{novo_misinfobot_2021}
T.~N.~F. Novo, ``{MisInfoBot}: fight misinformation about {COVID} on social media.''

\bibitem{bansal_deepfake_2023}
\BIBentryALTinterwordspacing
K.~Bansal, S.~Agarwal, and N.~Vyas, ``Deepfake detection using {CNN} and {DCGANS} to drop-out fake multimedia content: A hybrid approach,'' in \emph{2023 International Conference on {IoT}, Communication and Automation Technology ({ICICAT})}, pp. 1--6. [Online]. Available: \url{https://ieeexplore.ieee.org/document/10263628}
\BIBentrySTDinterwordspacing

\bibitem{botha_fake_2020}
\BIBentryALTinterwordspacing
J.~G. Botha and H.~Pieterse, \emph{Fake news and deepfakes: A dangerous threat for 21st century information security}, accepted: 2021-04-06T09:09:18Z Publication Title: Proceedings of the 15th International Conference on Cyber Warfare and Security, Norfolk, Virginia, 12-13 March 2020. [Online]. Available: \url{https://researchspace.csir.co.za/dspace/handle/10204/11946}
\BIBentrySTDinterwordspacing

\bibitem{wang_fakespotter_2020}
\BIBentryALTinterwordspacing
R.~Wang, F.~Juefei-Xu, L.~Ma, X.~Xie, Y.~Huang, J.~Wang, and Y.~Liu, ``{FakeSpotter}: A simple yet robust baseline for spotting {AI}-synthesized fake faces.'' [Online]. Available: \url{http://arxiv.org/abs/1909.06122}
\BIBentrySTDinterwordspacing

\bibitem{cardenuto_age_2023}
\BIBentryALTinterwordspacing
J.~P. Cardenuto, J.~Yang, R.~Padilha, R.~Wan, D.~Moreira, H.~Li, S.~Wang, F.~Andaló, S.~Marcel, and A.~Rocha, ``The age of synthetic realities: Challenges and opportunities.'' [Online]. Available: \url{http://arxiv.org/abs/2306.11503}
\BIBentrySTDinterwordspacing

\bibitem{mosallanezhad_topic-preserving_2020}
\BIBentryALTinterwordspacing
A.~Mosallanezhad, K.~Shu, and H.~Liu, ``Topic-preserving synthetic news generation: An adversarial deep reinforcement learning approach.'' [Online]. Available: \url{http://arxiv.org/abs/2010.16324}
\BIBentrySTDinterwordspacing

\bibitem{cocchi_unveiling_2023}
F.~Cocchi, L.~Baraldi, S.~Poppi, M.~Cornia, L.~Baraldi, and R.~Cucchiara, ``Unveiling the impact of image transformations on deepfake detection: An experimental analysis,'' in \emph{Image Analysis and Processing – {ICIAP} 2023}, ser. Lecture Notes in Computer Science, G.~L. Foresti, A.~Fusiello, and E.~Hancock, Eds.\hskip 1em plus 0.5em minus 0.4em\relax Springer Nature Switzerland, pp. 345--356.

\bibitem{allcott_social_2017}
\BIBentryALTinterwordspacing
H.~Allcott and M.~Gentzkow, ``Social media and fake news in the 2016 election.'' [Online]. Available: \url{https://papers.ssrn.com/abstract=2903810}
\BIBentrySTDinterwordspacing

\bibitem{lazer_science_2018}
\BIBentryALTinterwordspacing
D.~M.~J. Lazer, M.~A. Baum, Y.~Benkler, A.~J. Berinsky, K.~M. Greenhill, F.~Menczer, M.~J. Metzger, B.~Nyhan, G.~Pennycook, D.~Rothschild, M.~Schudson, S.~A. Sloman, C.~R. Sunstein, E.~A. Thorson, D.~J. Watts, and J.~L. Zittrain, ``The science of fake news,'' vol. 359, no. 6380, pp. 1094--1096. [Online]. Available: \url{http://arxiv.org/abs/2307.07903}
\BIBentrySTDinterwordspacing

\bibitem{zhou_network-based_2019}
\BIBentryALTinterwordspacing
X.~Zhou and R.~Zafarani, ``Network-based fake news detection: A pattern-driven approach.'' [Online]. Available: \url{http://arxiv.org/abs/1906.04210}
\BIBentrySTDinterwordspacing

\bibitem{bradshaw_industrialized_2021}
S.~Bradshaw, H.~Bailey, and P.~N. Howard, ``Industrialized disinformation: 2020 global inventory of organized social media manipulation (computational propaganda research project).''

\bibitem{shu_combating_2020}
K.~Shu, A.~Bhattacharjee, F.~Alatawi, T.~H.~Nazer, K.~Ding, M.~Karami, and H.~Liu, \emph{Combating Disinformation in a Social Media Age}.

\bibitem{vaswani_attention_2023}
\BIBentryALTinterwordspacing
A.~Vaswani, N.~Shazeer, N.~Parmar, J.~Uszkoreit, L.~Jones, A.~N. Gomez, L.~Kaiser, and I.~Polosukhin, ``Attention is all you need.'' [Online]. Available: \url{http://arxiv.org/abs/1706.03762}
\BIBentrySTDinterwordspacing

\bibitem{sandrini_generative_2023}
\BIBentryALTinterwordspacing
L.~Sandrini and R.~Somogyi, ``Generative {AI} and deceptive news consumption,'' vol. 232, p. 111317. [Online]. Available: \url{https://www.sciencedirect.com/science/article/pii/S0165176523003427}
\BIBentrySTDinterwordspacing

\bibitem{ferrara_genai_2024}
\BIBentryALTinterwordspacing
E.~Ferrara, ``{GenAI} against humanity: nefarious applications of generative artificial intelligence and large language models.'' [Online]. Available: \url{https://doi.org/10.1007/s42001-024-00250-1}
\BIBentrySTDinterwordspacing

\bibitem{goodfellow_generative_2014}
\BIBentryALTinterwordspacing
I.~J. Goodfellow, J.~Pouget-Abadie, M.~Mirza, B.~Xu, D.~Warde-Farley, S.~Ozair, A.~Courville, and Y.~Bengio, ``Generative adversarial networks.'' [Online]. Available: \url{http://arxiv.org/abs/1406.2661}
\BIBentrySTDinterwordspacing

\bibitem{kingma_auto-encoding_2022}
\BIBentryALTinterwordspacing
D.~P. Kingma and M.~Welling, ``Auto-encoding variational bayes.'' [Online]. Available: \url{http://arxiv.org/abs/1312.6114}
\BIBentrySTDinterwordspacing

\bibitem{ma_era_2024}
\BIBentryALTinterwordspacing
S.~Ma, H.~Wang, L.~Ma, L.~Wang, W.~Wang, S.~Huang, L.~Dong, R.~Wang, J.~Xue, and F.~Wei, ``The era of 1-bit {LLMs}: All large language models are in 1.58 bits.'' [Online]. Available: \url{http://arxiv.org/abs/2402.17764}
\BIBentrySTDinterwordspacing

\bibitem{radford_improving_2018}
A.~Radford, K.~Narasimhan, T.~Salimans, and I.~Sutskever, ``Improving language understanding by generative pre-training.''

\bibitem{del_bimbo_automatic_2021}
\BIBentryALTinterwordspacing
M.~Schütz, A.~Schindler, M.~Siegel, and K.~Nazemi, ``Automatic fake news detection with pre-trained transformer models,'' in \emph{Pattern Recognition. {ICPR} International Workshops and Challenges}, A.~Del~Bimbo, R.~Cucchiara, S.~Sclaroff, G.~M. Farinella, T.~Mei, M.~Bertini, H.~J. Escalante, and R.~Vezzani, Eds.\hskip 1em plus 0.5em minus 0.4em\relax Springer International Publishing, vol. 12667, pp. 627--641, series Title: Lecture Notes in Computer Science. [Online]. Available: \url{http://link.springer.com/10.1007/978-3-030-68787-8_45}
\BIBentrySTDinterwordspacing

\bibitem{loth_rise_2023-1}
\BIBentryALTinterwordspacing
A.~Loth, ``The rise of generative {AI}: Revolutionizing innovation and enhancing human collaboration.'' [Online]. Available: \url{https://www.researchgate.net/publication/369625534_The_Rise_of_Generative_AI_Revolutionizing_Innovation_and_Enhancing_Human_Collaboration}
\BIBentrySTDinterwordspacing

\bibitem{otis_true_2020}
C.~L. Otis, \emph{True Or False: A {CIA} Analyst's Guide to Spotting Fake News}.\hskip 1em plus 0.5em minus 0.4em\relax Feiwel \& Friends.

\bibitem{loth_decisively_2021}
A.~Loth, \emph{Decisively Digital: From Creating a Culture to Designing Strategy}.\hskip 1em plus 0.5em minus 0.4em\relax John Wiley \& Sons.

\bibitem{loth_ki_2024}
------, \emph{{KI} für Content Creation: Texte, Bilder, Audio erstellen mit {ChatGPT} \& Co.}\hskip 1em plus 0.5em minus 0.4em\relax {MITP}-Verlags {GmbH} \& Co. {KG}.

\bibitem{schutz_ait_fhstp_2022}
M.~Schütz, J.~Böck, M.~Andresel, A.~Kirchknopf, D.~Liakhovets, D.~Slijepčević, and A.~Schindler, ``{AIT}\_fhstp at {CheckThat}! 2022: Cross-lingual fake news detection with a large pre-trained transformer.''

\bibitem{loth_visual_2019}
A.~Loth, \emph{Visual analytics with Tableau}.\hskip 1em plus 0.5em minus 0.4em\relax John Wiley \& Sons.

\bibitem{peters_dissecting_2018}
M.~E. Peters, M.~Neumann, L.~Zettlemoyer, and W.-t. Yih, ``Dissecting contextual word embeddings: Architecture and representation.''

\bibitem{de_oliveira_identifying_2021}
\BIBentryALTinterwordspacing
N.~R. de~Oliveira, P.~S. Pisa, M.~A. Lopez, D.~S.~V. de~Medeiros, and D.~M.~F. Mattos, ``Identifying fake news on social networks based on natural language processing: Trends and challenges,'' vol.~12, no.~1, p.~38, number: 1 Publisher: Multidisciplinary Digital Publishing Institute. [Online]. Available: \url{https://www.mdpi.com/2078-2489/12/1/38}
\BIBentrySTDinterwordspacing

\bibitem{howard_universal_2018}
J.~Howard and S.~Ruder, ``Universal language model fine-tuning for text classification.''

\bibitem{vijjali_two_2020}
\BIBentryALTinterwordspacing
R.~Vijjali, P.~Potluri, S.~Kumar, and S.~Teki, ``Two stage transformer model for {COVID}-19 fake news detection and fact checking.'' [Online]. Available: \url{http://arxiv.org/abs/2011.13253}
\BIBentrySTDinterwordspacing

\bibitem{bubeck_sparks_2023}
\BIBentryALTinterwordspacing
S.~Bubeck, V.~Chandrasekaran, R.~Eldan, J.~Gehrke, E.~Horvitz, E.~Kamar, P.~Lee, Y.~T. Lee, Y.~Li, S.~Lundberg, H.~Nori, H.~Palangi, M.~T. Ribeiro, and Y.~Zhang, ``Sparks of artificial general intelligence: Early experiments with {GPT}-4.'' [Online]. Available: \url{http://arxiv.org/abs/2303.12712}
\BIBentrySTDinterwordspacing

\bibitem{zhavoronkov_artificial_2019}
\BIBentryALTinterwordspacing
A.~Zhavoronkov, P.~Mamoshina, Q.~Vanhaelen, M.~Scheibye-Knudsen, A.~Moskalev, and A.~Aliper, ``Artificial intelligence for aging and longevity research: Recent advances and perspectives,'' vol.~49, pp. 49--66. [Online]. Available: \url{https://linkinghub.elsevier.com/retrieve/pii/S156816371830240X}
\BIBentrySTDinterwordspacing

\bibitem{jin_generative_2020}
\BIBentryALTinterwordspacing
L.~Jin, F.~Tan, S.~Jiang, and R.~Köker, ``Generative adversarial network technologies and applications in computer vision,'' vol. 2020. [Online]. Available: \url{https://doi.org/10.1155/2020/1459107}
\BIBentrySTDinterwordspacing

\bibitem{arora_generative_2022}
\BIBentryALTinterwordspacing
A.~Arora and A.~Arora, ``Generative adversarial networks and synthetic patient data: current challenges and future perspectives,'' vol.~9, no.~2, pp. 190--193, publisher: Cambridge University Press. [Online]. Available: \url{https://www.ncbi.nlm.nih.gov/pmc/articles/PMC9345230/}
\BIBentrySTDinterwordspacing

\bibitem{radford_language_2019}
A.~Radford, J.~Wu, R.~Child, D.~Luan, D.~Amodei, and I.~Sutskever, ``Language models are unsupervised multitask learners.''

\bibitem{weisz_toward_2023}
\BIBentryALTinterwordspacing
J.~D. Weisz, M.~Muller, J.~He, and S.~Houde, ``Toward general design principles for generative {AI} applications.'' [Online]. Available: \url{http://arxiv.org/abs/2301.05578}
\BIBentrySTDinterwordspacing

\bibitem{simon_misinformation_2023}
\BIBentryALTinterwordspacing
F.~M. Simon, S.~Altay, and H.~Mercier, ``Misinformation reloaded? fears about the impact of generative {AI} on misinformation are overblown.'' [Online]. Available: \url{https://misinforeview.hks.harvard.edu/article/misinformation-reloaded-fears-about-the-impact-of-generative-ai-on-misinformation-are-overblown/}
\BIBentrySTDinterwordspacing

\bibitem{bradshaw_playing_2022}
\BIBentryALTinterwordspacing
S.~Bradshaw, R.~{DiResta}, and C.~Miller, ``Playing both sides: Russian state-backed media coverage of the \#{BlackLivesMatter} movement,'' p. 19401612221082052, publisher: {SAGE} Publications Inc. [Online]. Available: \url{https://doi.org/10.1177/19401612221082052}
\BIBentrySTDinterwordspacing

\bibitem{lewandowsky_countering_2021}
\BIBentryALTinterwordspacing
S.~Lewandowsky and S.~Van Der~Linden, ``Countering misinformation and fake news through inoculation and prebunking,'' vol.~32, no.~2, pp. 348--384. [Online]. Available: \url{https://www.tandfonline.com/doi/full/10.1080/10463283.2021.1876983}
\BIBentrySTDinterwordspacing

\bibitem{shu_fakenewsnet_2020}
K.~Shu, D.~Mahudeswaran, S.~Wang, D.~Lee, and H.~Liu, ``Fakenewsnet: A data repository with news content, social context, and spatiotemporal information for studying fake news on social media,'' vol.~8, no.~3, pp. 171--188, publisher: Mary Ann Liebert, Inc., publishers 140 Huguenot Street, 3rd Floor New ….

\bibitem{yang_others_2021}
\BIBentryALTinterwordspacing
J.~Yang and Y.~Tian, ``“others are more vulnerable to fake news than i am”: Third-person effect of {COVID}-19 fake news on social media users,'' vol. 125, p. 106950. [Online]. Available: \url{https://www.ncbi.nlm.nih.gov/pmc/articles/PMC8867061/}
\BIBentrySTDinterwordspacing

\bibitem{pennycook_psychology_2021}
\BIBentryALTinterwordspacing
G.~Pennycook and D.~G. Rand, ``The psychology of fake news,'' vol.~25, no.~5, pp. 388--402, publisher: Elsevier. [Online]. Available: \url{https://www.cell.com/trends/cognitive-sciences/abstract/S1364-6613(21)00051-6}
\BIBentrySTDinterwordspacing

\bibitem{verma2024one}
N.~Verma, ``“one video could start a war”: A qualitative interview study of public perceptions of deepfake technology,'' \emph{Proceedings of the Association for Information Science and Technology}, vol.~61, no.~1, pp. 374--385, 2024.

\bibitem{alsaid2024combating}
M.~Alsaid, S.~Parvathi~Panguluri, and S.~Hawamdeh, ``Combating misinformation on social media using social noise and social entropy as a measure of uncertainty,'' \emph{Proceedings of the Association for Information Science and Technology}, vol.~61, no.~1, pp. 25--35, 2024.

\bibitem{liu_early_2018}
\BIBentryALTinterwordspacing
Y.~Liu and Y.-F. Wu, ``Early detection of fake news on social media through propagation path classification with recurrent and convolutional networks,'' vol.~32, no.~1, number: 1. [Online]. Available: \url{https://ojs.aaai.org/index.php/AAAI/article/view/11268}
\BIBentrySTDinterwordspacing

\bibitem{przybyla_capturing_2020}
\BIBentryALTinterwordspacing
P.~Przybyla, ``Capturing the style of fake news,'' vol.~34, no.~1, pp. 490--497, number: 01. [Online]. Available: \url{https://ojs.aaai.org/index.php/AAAI/article/view/5386}
\BIBentrySTDinterwordspacing

\bibitem{qian_neural_2018}
\BIBentryALTinterwordspacing
F.~Qian, C.~Gong, K.~Sharma, and Y.~Liu, ``Neural user response generator: Fake news detection with collective user intelligence,'' in \emph{Proceedings of the Twenty-Seventh International Joint Conference on Artificial Intelligence}.\hskip 1em plus 0.5em minus 0.4em\relax International Joint Conferences on Artificial Intelligence Organization, pp. 3834--3840. [Online]. Available: \url{https://www.ijcai.org/proceedings/2018/533}
\BIBentrySTDinterwordspacing

\bibitem{shu_beyond_2019}
\BIBentryALTinterwordspacing
K.~Shu, S.~Wang, and H.~Liu, ``Beyond news contents: The role of social context for fake news detection,'' in \emph{Proceedings of the Twelfth {ACM} International Conference on Web Search and Data Mining}, ser. {WSDM} '19.\hskip 1em plus 0.5em minus 0.4em\relax Association for Computing Machinery, pp. 312--320. [Online]. Available: \url{https://dl.acm.org/doi/10.1145/3289600.3290994}
\BIBentrySTDinterwordspacing

\bibitem{silva_towards_2020}
\BIBentryALTinterwordspacing
R.~M. Silva, R.~L.~S. Santos, T.~A. Almeida, and T.~A.~S. Pardo, ``Towards automatically filtering fake news in portuguese,'' vol. 146, p. 113199. [Online]. Available: \url{https://www.sciencedirect.com/science/article/pii/S0957417420300257}
\BIBentrySTDinterwordspacing

\bibitem{goh2024humans}
D.~H.-L. Goh, J.~Pan, and C.~S. Lee, ``Humans versus machines: A deepfake detection faceoff,'' \emph{Proceedings of the Association for Information Science and Technology}, vol.~61, no.~1, pp. 917--919, 2024.

\bibitem{chun2024can}
R.~W.~Y. Chun, S.~Huang, D.~H.-L. Goh, C.~S. Lee, and Y.-L. Theng, ``Can seniors spot deepfakes? a diary study of deepfake identification strategies,'' \emph{Proceedings of the Association for Information Science and Technology}, vol.~61, no.~1, pp. 877--879, 2024.

\bibitem{pennycook_who_2020}
\BIBentryALTinterwordspacing
G.~Pennycook and D.~G. Rand, ``Who falls for fake news? the roles of bullshit receptivity, overclaiming, familiarity, and analytic thinking,'' vol.~88, no.~2, pp. 185--200, \_eprint: https://onlinelibrary.wiley.com/doi/pdf/10.1111/jopy.12476. [Online]. Available: \url{https://onlinelibrary.wiley.com/doi/abs/10.1111/jopy.12476}
\BIBentrySTDinterwordspacing

\bibitem{tandoc_jr_defining_2018}
E.~C. Tandoc~Jr, Z.~W. Lim, and R.~Ling, ``Defining “fake news” a typology of scholarly definitions,'' vol.~6, no.~2, pp. 137--153, publisher: Taylor \& Francis.

\bibitem{egelhofer_fake_2019}
J.~L. Egelhofer and S.~Lecheler, ``Fake news as a two-dimensional phenomenon: A framework and research agenda,'' vol.~43, no.~2, pp. 97--116, publisher: Taylor \& Francis.

\bibitem{osmundsen_partisan_2021}
\BIBentryALTinterwordspacing
M.~Osmundsen, A.~Bor, P.~B. Vahlstrup, A.~Bechmann, and M.~B. Petersen, ``Partisan polarization is the primary psychological motivation behind political fake news sharing on twitter,'' vol. 115, no.~3, pp. 999--1015, publisher: Cambridge University Press. [Online]. Available: \url{https://www.cambridge.org/core/journals/american-political-science-review/article/abs/partisan-polarization-is-the-primary-psychological-motivation-behind-political-fake-news-sharing-on-twitter/3F7D2098CD87AE5501F7AD4A7FA83602}
\BIBentrySTDinterwordspacing

\bibitem{noauthor_podexcellence_2023}
\BIBentryALTinterwordspacing
Pôle d’excellence cyber: Lutte contre les manipulations de l’information avril 2023. [Online]. Available: \url{https://www.pole-excellence-cyber.org/wp-content/uploads/2024/01/LMI_PEC_2023.pdf}
\BIBentrySTDinterwordspacing

\bibitem{noauthor_podexcellence_2024}
\BIBentryALTinterwordspacing
Pôle d’excellence cyber: Lutte contre les manipulations de l’information, regards croisés de spécialistes et d’acteurs du domaine tome 2. [Online]. Available: \url{https://www.pole-excellence-cyber.org/wp-content/uploads/2024/11/LMI_Tome2_PEC2024.pdf}
\BIBentrySTDinterwordspacing

\end{thebibliography}

\end{document}